\documentclass[lettersize,journal]{IEEEtran}
\usepackage{amsmath,amsfonts}
\usepackage{algorithmic}
\usepackage{algorithm}
\usepackage{array}
\usepackage[caption=false,font=normalsize,labelfont=sf,textfont=sf]{subfig}
\usepackage{textcomp}
\usepackage{stfloats}
\usepackage{url}
\usepackage{verbatim}
\usepackage{graphicx}
\usepackage{cite}
\usepackage{multirow}
\usepackage{mathrsfs}
\usepackage{times}
\usepackage{epsfig}
\usepackage{amssymb}
\usepackage{mathtools}
\usepackage{setspace}
\usepackage{booktabs}
\usepackage{color,xcolor}
\usepackage{bbding}
\usepackage{diagbox}
\usepackage{bm}
\usepackage{setspace}

\begin{document}

\title{Point Tree Transformer for Point Cloud Registration}

\author{Meiling Wang$^1$, Guangyan Chen$^1$, Yi Yang$^1$, Li Yuan$^2$, Yufeng Yue$^{1*}$
\thanks{This work is partly supported by the National Natural Science Foundation of China under Grant  62003039, 62233002, 61973034, the CAST program under Grant No. YESS20200126. (Corresponding Author: Yufeng Yue, yueyufeng@bit.edu.cn)}
\thanks{$^1$Meiling Wang, Guangyan Chen, Yi Yang, and Yufeng Yue are with the School of Automation, Beijing Institute of Technology, Beijing, 100081, China.}
\thanks{$^2$Li Yuan is with the School of Electrical and Computer Engineering at Peking University; Pecheng Lab; Shenzhen, 518055, China. }

}

\maketitle

\begin{abstract}
Point cloud registration is a fundamental task in the fields of computer vision and robotics. Recent developments in transformer-based methods have demonstrated enhanced performance in this domain. However, the standard attention mechanism utilized in these methods often integrates many low-relevance points, thereby struggling to prioritize its attention weights on sparse yet meaningful points. This inefficiency leads to limited local structure modeling capabilities and {quadratic computational complexity}. To overcome these limitations, we propose the Point Tree Transformer (PTT), a novel transformer-based approach for point cloud registration that efficiently extracts comprehensive local and global features while maintaining linear computational complexity. The PTT constructs hierarchical feature trees from point clouds in a coarse-to-dense manner, and introduces a novel Point Tree Attention (PTA) mechanism, which follows the tree structure to facilitate the progressive convergence of attended regions towards salient points. Specifically, each tree layer selectively identifies a subset of key points with the highest attention scores. Subsequent layers focus attention on areas of significant relevance, derived from the child points of the selected point set. The feature extraction process additionally incorporates coarse point features that capture high-level semantic information, thus facilitating local structure modeling and the progressive integration of multiscale information. Consequently, PTA empowers the model to concentrate on crucial local structures and derive detailed local information while maintaining linear computational complexity. Extensive experiments conducted on the 3DMatch, ModelNet40, and KITTI datasets demonstrate that our method achieves superior performance over the state-of-the-art methods. 
\end{abstract}

\begin{IEEEkeywords}
Point cloud registration, Deep learning, Efficient Transformer, Transformer-based methods
\end{IEEEkeywords}

\begin{figure}[!t]
\setlength{\abovecaptionskip}{0pt}
    \centering
    \includegraphics[width=8.5cm]{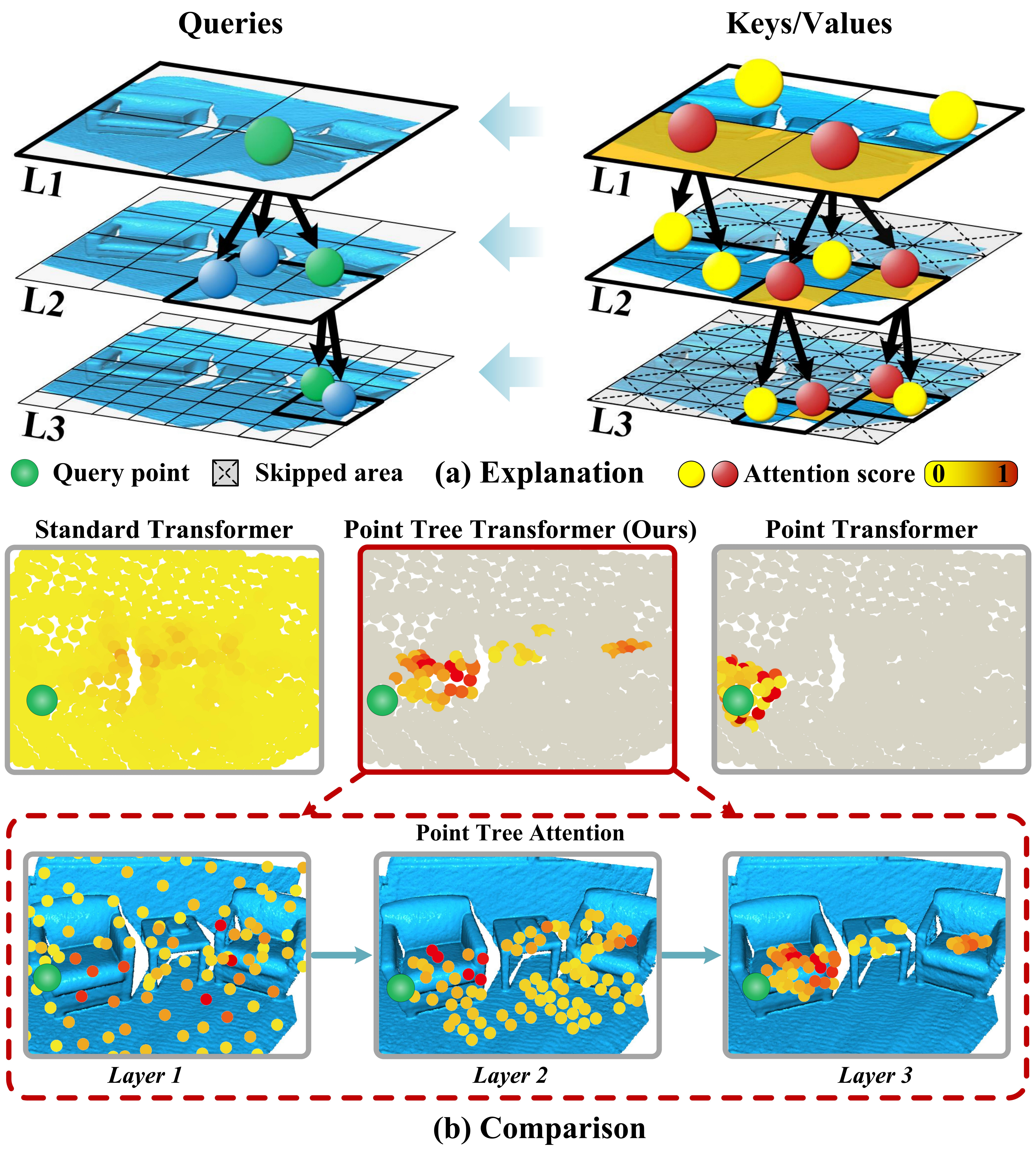}
    \caption{{Explanation of the Point Tree Attention (PTA) and \textcolor{black}{a comparison to the attention mechanisms in the Standard Transformer (ST) \cite{vaswani2017attention} and Point Transformer (PT) \cite{zhao2021point}, with visualization of attention weights for the point marked with a green dot.}} (a) {In our method, feature trees are built, and then PTA is used to hierarchically incorporate coarse features and restrict the attended regions of the next layer to the child points of the top $\mathcal{S}$ keys with the highest attention scores, skipping the shaded regions, where the locations of the top $\mathcal{S}$ keys are highlighted in the same color as the query}. Therefore, (b) PTA can focus on critical local structures and adaptively attend to relevant regions.  \textcolor{black}{In comparison, ST considers many low-relevance points and struggles to capture the local structures, whereas PT simply sets attention regions to predefined templates, overlooking information from other relevant regions.}}
    \label{fig1}
    \vspace{-12pt}
    \end{figure}

\section{Introduction}

Point cloud registration aims to determine an optimal transformation for aligning point cloud pairs, which is a fundamental problem in computer vision and robotics, such as 3D object detection, simultaneous localization and mapping (SLAM). 
{Recent advances in 3D point representation learning are pushing point cloud registration beyond the traditional methods \cite{besl1992method,segal2009generalized,zhou2016fast} to learning-based methods \cite{choy2020deep,li2021pointnetlk,aoki2019pointnetlk,wang2019deep, wu2022inenet, wu2023sacf, ren2022corri2p,huang2017coarse,zhang2022vrnet,wang2024neighborhood,cao2024sharpgconv,yuan2024learning,jiang2024gtinet,wang2023global}.} The most widely known traditional method is the iterative closest point (ICP) \cite{besl1992method},
which iterates between establishing correspondences and calculating a transformation.
However, ICP and its variants \cite{segal2009generalized,rusinkiewicz2001efficient} are prone to becoming stuck in local minima when the initial errors are large. 
To achieve increased registration accuracy,  learning-based methods \cite{choy2020deep,li2021pointnetlk,aoki2019pointnetlk} integrate neural networks to extract pointwise features separately and establish correspondences based on feature similarity.  However, the independence across point clouds produces obstacles when identifying common
structures and extracting distinct features. 

Learning-based methods \cite{shi2021keypoint,wang2019deep,wang2019prnet,fu2021robust} have endeavored to tackle these challenges by integrating transformer models, renowned for their adeptness in handling permutation invariance and capturing dependencies. These approaches enable one point cloud to perceive another point cloud and extract the contextual information between them, thereby augmenting the discriminative efficacy of the extracted features. Nonetheless, the standard attention mechanism often integrates many low-relevance points, which impedes its ability to effectively assign attention weights to sparse but significant points. This inefficiency leads to limited local structure modeling capabilities and imposes quadratic computational complexity.

{Many recent investigations \cite{zhao2021point, lu2022transformers, lai2022stratified, cheng2021patchformer, he2022voxel, mao2021voxel} have delved into local attention mechanisms for point cloud processing, which prune low-relevance points by confining attention to static, predefined patterns. \textcolor{black}{Point Transformer \cite{zhao2021point} constrains the attention scope within the local neighborhood, while VoTr \cite{mao2021voxel} modifies the attention framework by introducing local windows and stride dilation. However, these strategies limit their capability to dynamically prioritize highly relevant regions.} Furthermore, these methods presuppose correlations based on proximity between points, e.g., spatially close points are correlated. Such assumptions are typically inaccurate in cross-point-cloud scenarios, thereby complicating the efficacy of fixed attention patterns in cross-attention mechanisms. 
Consequently, there remains an essential requirement in point cloud registration to develop a transformer model capable of encoding crucial local structures while reducing computational complexity.}

 {To achieve this goal, we propose the Point Tree Transformer (PTT), which is capable of focusing on important local structures and achieving linear computational complexity without predefined attention sparsity.} The PTT is built on the basis of a proposed Point Tree Attention (PTA) module, 
which drives the hierarchical convergence of the attended regions and incorporates the spatially coarse features into the child points to guide the feature extraction process. \textcolor{black}{An intuitive example is visualized in Fig.\hspace{4pt}\ref{fig1}(a).  In the 1st layer, the attention computation for the query point encompasses all key points, from which the top $\mathcal{S}$ (here, $\mathcal{S}=2$) points with the highest attention scores, highlighted in orange, are selected. In the 2nd layer, for the child points corresponding to the query point in the previous layer (1st), attention is calculated exclusively among the child points of the corresponding $\mathcal{S}$ keys selected in the previous layer, thereby skipping low-relevance points and reducing the computational complexity. Furthermore, the features derived from the previous layer are utilized to guide the feature extraction procedure for the child points.
These processes are replicated in the 3rd layer, using the top $\mathcal{S}$ points selected in the 2nd layer. In this manner, PTA enables our method to adaptively specify high-relevance locations as attended areas and focus on critical local structures}. \textcolor{black}{Furthermore, the dynamic attention sparsity of PTA obviates the need for predefined patterns and is inherently compatible with cross-attention mechanisms. Experiments conducted on the 3DMatch, ModelNet40, and KITTI datasets demonstrate that our PTA mechanism efficiently extracts critical local structures while maintaining linear computational complexity. Consequently, our PTT accurately and efficiently aligns point clouds, outperforming state-of-the-art (SOTA) methods.} The main contributions are three-fold:

\begin{itemize}
\setlength{\itemsep}{5pt}
\setlength{\parsep}{5pt}
\setlength{\parskip}{5pt}
  \item  The PTT is proposed by integrating tree structures into the transformer model, allowing the model to extract rich local features and achieve linear computational complexity with the learned attended regions.
  \item  \textcolor{black}{The PTA is proposed to hierarchically and dynamically specify high-relevance key points and structurize point clouds along the tree, facilitating local structure modeling and multiscale information aggregation.}
  \item Extensive experiments show that our method outperforms the baselines and achieves SOTA performance on the 3DMatch, ModelNet40, and KITTI datasets.
\end{itemize}

\section{Related Work}
\subsection{{Transformer-based Methods for Registration}}
Inspired by the success of transformers in natural language processing (NLP) \cite{devlin2018bert,brown2020language,krizhevsky2017imagenet} and computer vision tasks \cite{liu2021swin,yuan2021tokens,yuan2021volo, tang2022quadtree}, researchers have adapted them to point cloud registration. 
The deep closest point (DCP) \cite{wang2019deep} utilizes a dynamic graph CNN (DGCNN) \cite{phan2018dgcnn} to separately extract features and introduces a transformer \cite{vaswani2017attention} to model relations across a pair of point clouds. 
The robust graph matching (RGM) method \cite{fu2021robust} adopts a transformer to improve the quality of the correspondences by aggregating information along graph edges. Predator \cite{huang2021predator} leverages self- and cross-attention mechanisms to perform information aggregation across a pair of point clouds and predict overlapping regions for feature sampling, which significantly improves the proportion of successful registrations in low-overlap scenarios. CoFiNet \cite{yu2021cofinet} alternately uses self-attention and cross-attention to extract features in a coarse-to-fine manner and achieves promising performance.
The registration transformer  (RegTR) \cite{yew2022regtr} utilizes attention layers to directly generate correspondences without nearest neighbor feature matching or RANSAC. {Geometric transformer \cite{qin2022geometric} calculates pair-wise distances and triplet-wise angles, which are then combined with self-attention to capture geometric features, ultimately leading to robust superpoint matching.}  DIT \cite{chen2023full} introduces a full Transformer network that leverages the transformer architecture to extract local features and facilitate deep information interaction. RegFormer \cite{liu2023regformer} presents an end-to-end transformer network designed for large-scale point cloud alignment. It achieves competitive performance in accuracy and efficiency, all without the need for additional post-processing steps. {In general, most current transformer-based methods utilize attention mechanisms for contextual information learning.} However, the standard attention mechanism struggles to focus its attention weights on meaningful points, leading to limited local structure modeling ability.

\subsection{{Local Attention for Point Clouds}}
To assist transformers in focusing on meaningful points and enhance their local feature extraction capabilities,
local attention mechanisms \cite{lu2022transformers, lai2022stratified, cheng2021patchformer, he2022voxel, mao2021voxel} introduce a local inductive bias by restricting the attention fields to local regions instead of the entire point cloud.
The sparse voxel transformer \cite{fan2022svt} encodes short-range local relations based on voxels and learns long-range contextual relations based on clusters.
The point transformer \cite{zhao2021point} applies an attention mechanism in the local neighborhood of the given points.
The voxel transformer \cite{mao2021voxel}  converges attended regions to their local vicinity to prune low-relevance points while sampling distant points to achieve large receptive fields. 
PatchFormer \cite{cheng2021patchformer} splits a raw point cloud into M patches, aggregates the local features in each patch, and then approximates the global attention map.
The voxel set transformer \cite{he2022voxel} formulates the self-attention in each voxel by means of two cross-attention modules and models features in a set-to-set fashion.
The stratified transformer \cite{lai2022stratified}  captures
the short-range dependencies within a voxel and models long-range relations with respect to the downsampled point cloud outside the voxel. In summary, most existing local attention mechanisms for point clouds limit the attention field by considering a fixed pattern. {This makes it difficult to accurately attend to high-relevance points and achieve cross-attention, which is important for registration.}

\section{Point Tree Transformer}

\begin{figure*}[ht]
 \setlength{\abovecaptionskip}{-0.13cm}
    \begin{center}
    \includegraphics[width=18.0cm]{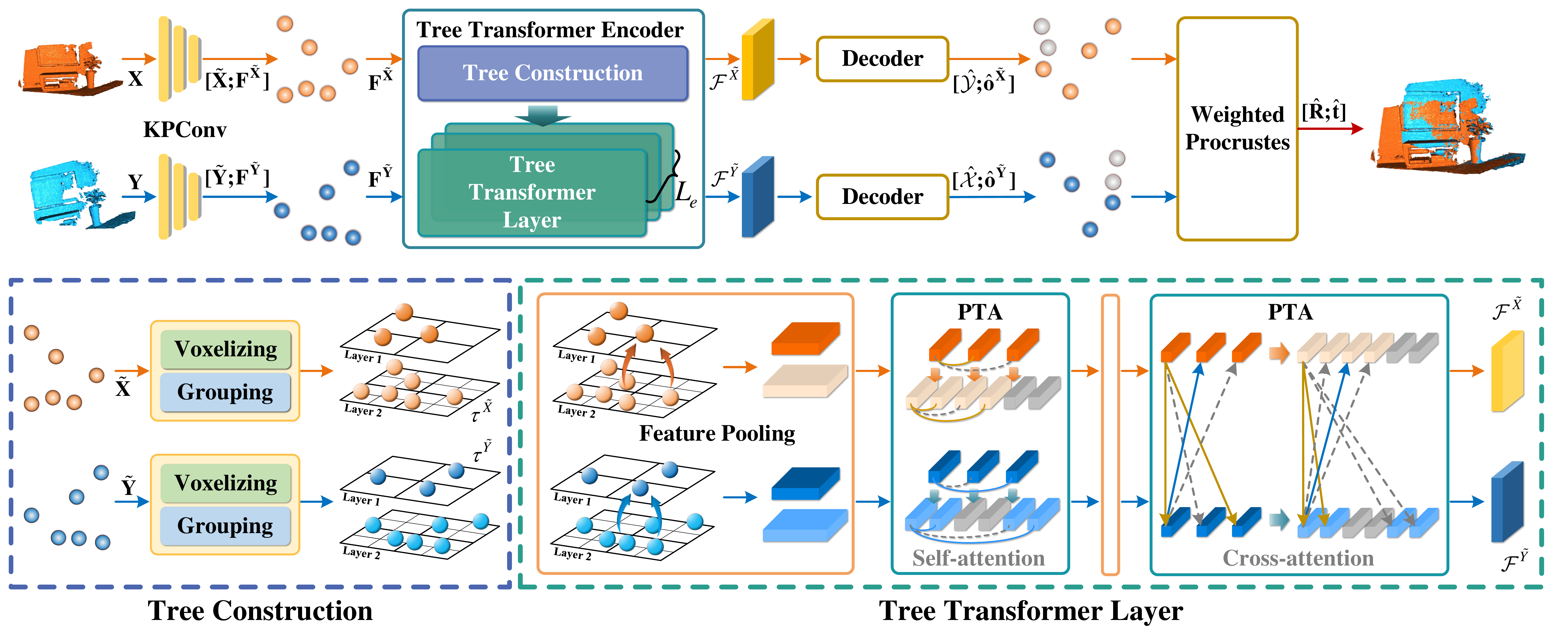}
    \end{center}
       \caption{{ Network architecture of the PTT.} First, the KPConv extracts features for a sparse set of points. Subsequently, the tree transformer encoder builds feature trees and iteratively extracts features containing local and global information. Then, the decoder predicts the corresponding point clouds and overlap scores. Finally, a transformation is computed to align the point clouds.}
    \label{Overview}
\end{figure*}

\subsection{Preliminaries}
{Transformers have demonstrated promising results in point clouds due to their advantages of order-invariance and dependency modeling provided by attention mechanisms. 
The standard transformer consists of a multilayer perceptron ($\rm{ MLP}$), layer normalization ($\rm{  LN}$), and a multihead attention operation ($\rm{  MA}$) that executes $H$ attention functions $\rm{Att}$ in parallel.
Given a pair of point cloud embeddings ${{\bm F}^{\tilde{X}}}$ and ${{\bm F}^{\tilde{Y}}}$ as inputs to $\rm{  MA}$, each $\rm{Att}$ generates queries $\bm{Q}$, keys $\bm{K}$, and values $\bm V$, using projection matrices ${\bm W}^Q$, ${\bm W}^K$, and ${\bm W}^V$, respectively:
\begin{equation} \label{Projection} \small
\setlength{\abovedisplayskip}{3pt}
\setlength{\belowdisplayskip}{3pt}
\begin{split}
    \bm{Q} =  {\bm F}^{\tilde{X}} {\bm W}^Q, \quad {\bm K} =  {\bm F}^{\tilde{Y}} {\bm W}^K, \quad {\bm V} = {\bm F}^{\tilde{Y}} {\bm W}^V.
\end{split}
\end{equation}
Then, each $\rm Att$ obtains an attention map via a scaled dot-product operation and multiplies this map by $V$ for information aggregation. Subsequently, the results are concatenated and projected with ${\bm W}^O$ to obtain the final values:
\begin{equation} \small
\label{attention}
\setlength{\abovedisplayskip}{3pt}
\setlength{\belowdisplayskip}{3pt}
\begin{split} 
    {\bm  A} &= {\rm softmax}(\frac{\bm{ QK}^T}{\sqrt{d_K}}){\bm V}, \\
    {\rm  MA}({\bm F}^{\tilde{X}},{\bm F}^{\tilde{Y}}) &= {  {\rm Concat}}({\bm A}_1, \dots, {\bm A}_{H}){\bm W}^O, \\
\end{split}
\end{equation}
where $d_K$ is the dimensionality of the keys $\bm K$.
The attention mechanism establishes associations across $\bm F^{\tilde{X}}$ and $ \bm  F^{\tilde{Y}}$, enabling $\bm F^{\tilde{X}}$ to receive information from $\bm F^{\tilde{Y}}$. Despite its versatile and powerful relation modeling capabilities, the standard attention suffers from a limited local feature extraction ability and quadratic computational complexity.  

\subsection{Overall Architecture}
The objective of point cloud registration is to estimate a rotation matrix $ \hat{\bm R} \in  SO(3)$ and a translation vector $\hat{\bm t} \in \mathbb{R}^3$ to align a source point cloud $\bm X =\left\{{x_{1},x_{2},...,x_{M}}\right\}  \subseteq \mathbb{R}^{3}$  with a target point cloud $\bm Y = \left\{{y_{1},y_{2},...,y_{N}}\right\}  \subseteq \mathbb{R}^{3}$. 

The overall PTT pipeline is illustrated in Fig.\hspace{4pt}\ref{Overview}. The PTT begins with a kernel point convolution (KPConv) \cite{thomas2019kpconv} (Sec.\hspace{4pt}\ref{KPConv}) that downsamples point clouds $\bm X, \,\bm Y$ into smaller sets of points ${\tilde{{\bm X}}}, \,  \tilde{{\bm Y}}$ and extracts pointwise features $ {\bm F}^{\tilde{X}}, \,  {\bm F}^{\tilde{Y}}$. Subsequently, a tree transformer encoder learns contextual information and extracts features $ \bm{\mathcal{{F}}}^{\tilde{X}}, \,  { \bm{\mathcal{{F}}}^{\tilde{Y}}}$ with rich local information (Sec.\hspace{4pt}\ref{Tree}). These features are then utilized to generate the corresponding point clouds $\hat{{\bm Y}}, \, \hat{{\bm X}}$ and predict overlap scores $\hat{{\bm o}}^{\tilde{X}}, \, \hat{{\bm o}}^{\tilde{Y}}$ in the decoder (Sec.\hspace{4pt}\ref{Decoder}). Finally, a weighted Procrustes module estimates the optimal transformation $\{ \hat{{\bm R}},\hat{{\bm t}}\} $ based on the predicted correspondences $\{\tilde{{\bm X}},\hat{{\bm Y}}\}, \, \{\tilde{{\bm Y}}, \hat{{\bm X}}\}$ and the overlap scores $\hat{{\bm o}}^{\tilde{X}}, \, \hat{{\bm o}}^{\tilde{Y}}$. The notations utilized in this article are summarized in Appendix G for convenience.

\subsection{Downsampling and Feature Extraction}\label{KPConv}
{Following \cite{huang2021predator}, a KPConv backbone, which consists of ResNet-like blocks and strided convolutions, is utilized for downsampling and feature extraction}. Specifically, the KPConv backbone downsamples the point clouds ${\bm X}  \in \mathbb{R}^{M \times 3}, \, {\bm Y}  \in \mathbb{R}^{N \times 3}$ to $\tilde{{\bm X}}  \in \mathbb{R}^{M^{'} \times 3}, \, \tilde{{\bm Y}}  \in \mathbb{R}^{N^{'} \times 3}$ and performs feature extraction. The extracted features are then projected to obtain features ${\bm F}^{\tilde{X} }  \in \mathbb{R}^{M^{'} \times D}, \, {\bm F}^{\tilde{Y} }  \in \mathbb{R}^{N^{'} \times D}$.

\subsection{Tree Transformer Encoder} \label{Tree}
The downsampled point clouds $\tilde{{\bm X}}, \, \tilde{{\bm Y}}$ and the extracted features ${\bm F}^{\tilde{X} }, \, {\bm F}^{\tilde{Y} }$ are processed by the tree transformer encoder. The encoder is composed of a tree construction layer, which generates trees ${\bm \tau}^{\tilde{X}}, \, {\bm \tau}^{\tilde{Y}}$ to represent the point clouds, as well as $L_{e}$ encoder layers for feature extraction. Each encoder layer consists of two feature pooling sub-layers and two PTA sub-layers, extracting features in linear computational complexity.

\textbf{Tree construction}. 
To encourage the attention weights to converge toward meaningful points and progressively structurize point clouds, tree structures are employed to represent point clouds. The intuitive procedure of constructing a 3-layer tree representation is shown in Fig. \ref{Tree_cons}.
Specifically, $L_{\tau}$-layer trees ${\bm \tau}^{\tilde{X}}$ and ${\bm \tau}^{\tilde{Y}}$ are built upon point clouds $\tilde{{\bm X}} $ and $ \tilde{{\bm Y}} $, respectively, by first dividing the point clouds into voxels and then hierarchically grouping $\mathbb{N}$ adjacent voxels into one voxel. The constructed tree ${\bm \tau}^{\tilde{X}}$ is defined by (1) the respective sets ${\bm P}_{l}^{\tilde{X}}$ of $N_{l}^{\tilde{X}}$ points in different layers $l = 1,2,...,L_{\tau}$; {(2) the coarse-to-dense indices ${\bm \rho}_{c \rightarrow d}^{\tilde{X}}$, which denote the indices of the child points corresponding to the $i$th coarse point; and (3) the dense-to-coarse indices ${\bm \rho}_{d \rightarrow c}^{\tilde{X}}$, which denote the parent point index of the $i$th dense point, with $c\! =\! 1,2,...,L_{\tau}\!-\!1$ and $ d\!=\!c\!+\!1$.}
The coordinates ${\bm C}_{c}^{\tilde{X}} \in\! \mathbb{R}^{N_{c}^{\tilde{X}} \times 3}$ of the coarse points are obtained by averaging the coordinates ${\bm C}_{d}^{\tilde{X}}$ of the child points. Specifically, the coordinates ${\bm C}_{c}^{\tilde{X}i}$ of the $i$th coarse point ${\bm P}_{c}^{\tilde{X}i}$ are obtained as follows:
\begin{equation} \label{Projection} \small
\setlength{\abovedisplayskip}{3pt}
\setlength{\belowdisplayskip}{3pt}
\begin{split}
    {\bm C}_{c}^{\tilde{X}i} = \frac{1}{{|{\bm \rho}_{c \!\rightarrow \! d}^{\tilde{X}i}|}} \sum_{j \in {\bm \rho}_{c \!\rightarrow \!d}^{\tilde{X}i}} {\bm C}_{d}^{\tilde{X}j},
\end{split}
\end{equation}
where ${\bm C}_{L_{\tau}}^{\tilde{X}}\!=\!\tilde{X}$ and $|{\bm \rho}_{c \rightarrow d}^{\tilde{X}i}|$ is the cardinality of ${\bm \rho}_{c \rightarrow d}^{\tilde{X}i}$.

\begin{figure}[t!]
    \begin{center}
    \includegraphics[width=8.5cm]{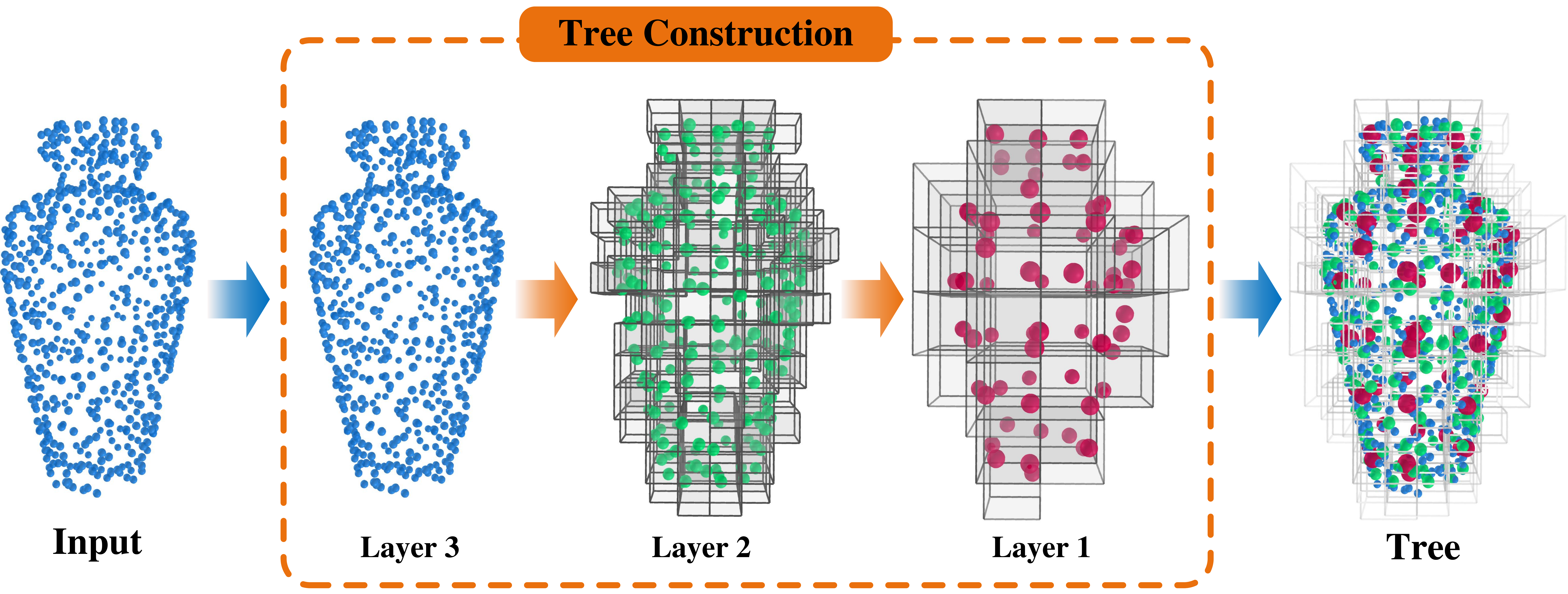}
    \end{center}
    \setlength{\abovecaptionskip}{-0.15cm}
   \caption{Illustration of tree construction. The point cloud in the densest layer utilizes the input point cloud and the sub-dense point cloud is obtained by voxelizing the densest point cloud. Then, the following layers group $\mathbb{N}$ voxels into one voxel to obtain the point cloud.}
    \label{Tree_cons}
\end{figure}

\textbf{Feature pooling}. {To aggregate information and build feature trees ${\bm F}_{l}^{\tilde{X}}\! \in\! \mathbb{R}^{N_{l}^{\tilde{X}} \times D}$ and ${\bm F}_{l}^{\tilde{Y}} \!\in \!\mathbb{R}^{N_{l}^{\tilde{Y}} \times D}$ ($l = 1,2,...,L_{\tau}$), the input features are initially utilized in the densest layer, and then the features are hierarchically aggregated to the corresponding parent points in a dense-to-coarse manner}. We hypothesize that the contributions of child points to their parent points are related to their relative positions. To facilitate an adaptive recalibration of the pointwise features in accordance with their respective contributions, {the dense features are concatenated along with their relative positions} and then projected using a two-layer ${\rm MLP}$, which comprises two fully connected layers and rectified linear unit (${\rm ReLU}$) activation. Concretely, the features ${\bm F}_{c}^{\tilde{X}i}$ for the $i$th coarse point ${\bm P}_{c}^{\tilde{X}i}$ are given as follows:
\begin{equation} \small
\setlength{\abovedisplayskip}{3pt}
\setlength{\belowdisplayskip}{5pt}
\begin{split}
                {\bm F}_{c}^{\tilde{X}i} = \frac{1}{{|{\bm \rho}_{c \!\rightarrow \! d}^{\tilde{X}i}|}} \sum_{j \in {\bm \rho}_{c \!\rightarrow \!d}^{\tilde{X}i}} {\rm MLP}({\rm Concat}({\bm F}_{d}^{\tilde{X}j},{\bm C}_{d}^{\tilde{X}j}\!-\! {\bm C}_{c}^{\tilde{X}i})).
\end{split}
\end{equation}

{By aggregating information, the feature trees ${\bm F}_{l}^{\tilde{X}}$ and $  {\bm F}_{l}^{\tilde{Y}}$ ($l\!=\! 1,2,...,L_{\tau}$) are built, where the level of semantic information increases from dense to coarse.}

    \textbf{Positional encoding}. {
    To succinctly and efficiently integrate positional information into each layer of the feature tree, we capitalize on the fact that both the coordinates and features of the coarse points are aggregated from the dense points through analogous procedures. Consequently, sinusoidal positional encoding \cite{vaswani2017attention} is directly integrated into the densest features prior to the feature pooling process.}
            
\begin{figure}[t] 
    \centering
    \includegraphics[width=8.5cm]{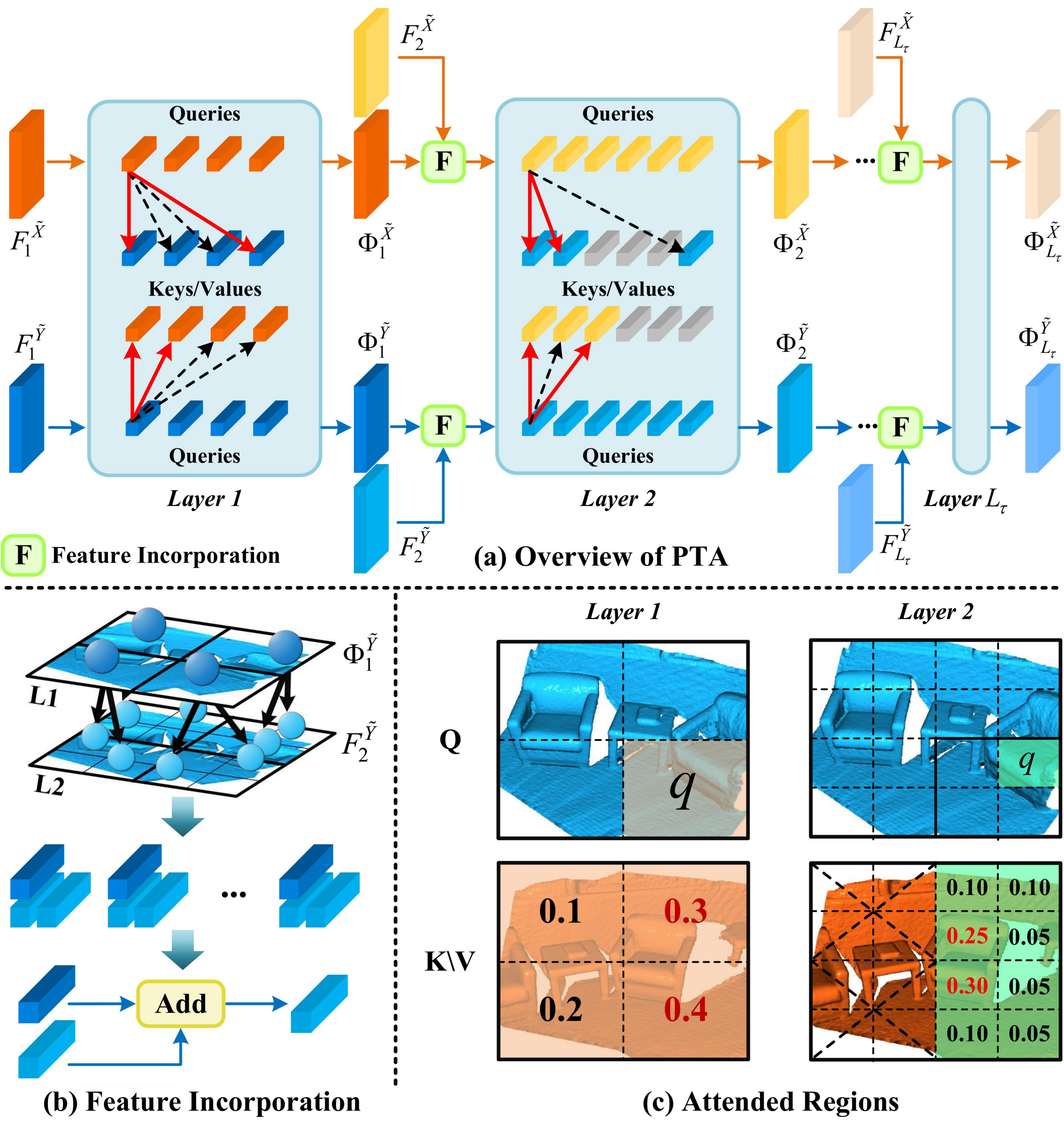}
    \caption{(a) Illustration of the PTA module, where \textbf{\textcolor{red}{$\uparrow $}} indicates the selected top $\mathcal{S}$ key points. (b) PTA incorporates spatially coarse features into the corresponding child points to guide the feature extraction process. {(c) Additionally, PTA hierarchically selects the top $\mathcal{S}$ key points with the highest attention scores. In each layer beyond the first, attention is evaluated only in the specified regions, which correspond to the child points of the $\mathcal{S}$ selected coarse key points. These specified regions are highlighted in the same color as the query}.}
    \label{TreeAttention}
        \end{figure}

\textbf{Point Tree Attention}. To capture important local structures and reduce computational complexity, PTA is introduced to progressively specify the attended regions and structurize the point clouds.
Consider a general case in which feature trees ${\bm F}_{l}^{\tilde{X}} $ and $  {\bm F}_{l}^{\tilde{Y}}$ of two different point clouds are given, where $l = 1,2,...,L_{\tau}$.
As shown in Fig.\hspace{4pt}\ref{TreeAttention}, PTA starts from the coarsest layer performs global attention ${\rm MA}({\bm F}_{1}^{\tilde{X}},{\bm F}_{1}^{\tilde{Y}}), \, {\rm MA}({\bm F}_{1}^{\tilde{Y}},{\bm F}_{1}^{\tilde{X}})$ to obtain the averaged attention maps $\mathbb{M}_{1}^{\tilde{X}} \in \mathbb{R}^{N_{1}^{\tilde{X}} \times N_{1}^{\tilde{Y}}}, \, \mathbb{M}_{1}^{\tilde{Y}} \in \mathbb{R}^{N_{1}^{\tilde{Y}} \times N_{1}^{\tilde{X}}}$ and the extracted features $\bm{\Phi}_{1}^{\tilde{X}} \in \mathbb{R}^{N_{1}^{\tilde{X}} \times D}, \, \bm{\Phi}_{1}^{\tilde{Y}} \in \mathbb{R}^{N_{1}^{\tilde{Y}} \times D}$. {In the next layer, PTA incorporates the extracted features $\bm{\Phi}_{1}^{\tilde{X}}, \, \bm{\Phi}_{1}^{\tilde{Y}}$ to guide the feature extraction process}, and specifies the attended regions based on the attention maps $\mathbb{M}_{1}^{\tilde{X}}, \, \mathbb{M}_{1}^{\tilde{Y}}$, then computes the attention within the specified regions. This process is iteratively repeated using shared parameters, until the densest layer is reached, obtaining the features $\bm{\Phi}_{L_{\tau}}^{\tilde{X}}$ and $\bm{\Phi}_{L_{\tau}}^{\tilde{Y}}$ that are enriched with mutual information. 

To better illustrate PTA, we consider any two consecutive layers, referred to as the coarse layer $c$ and the dense layer $d$ ($d=c+1$). Given the averaged attention maps $\mathbb{M}_{c}^{\tilde{X}}\!\in\! \mathbb{R}^{N_{c}^{\tilde{X}} \times \mathcal{K}_{c}^{{\tilde{Y}}}}, \mathbb{M}_{c}^{\tilde{Y}}\!\in \!\mathbb{R}^{N_{c}^{\tilde{Y}} \times \mathcal{K}_{c}^{{\tilde{X}}}}$ and the extracted features $\bm{\Phi}_{c}^{\tilde{X}}\! \in \!\mathbb{R}^{N_{c}^{\tilde{X}} \times D}, \, \bm{\Phi}_{c}^{\tilde{Y}}\! \in\! \mathbb{R}^{N_{c}^{\tilde{Y}} \times D}$ from the coarse layer, where $\mathcal{K}$ denotes the number of keys to which each query attends, with $\mathcal{K}_{1}^{{\tilde{Y}}}\!=\!N_1^{\tilde{Y}}, \, \mathcal{K}_{1}^{{\tilde{X}}}\!=\!N_1^{\tilde{X}}$. The procedure in the dense layer is detailed. Initially, to facilitate the local feature extraction and  multiscale information aggregation, the extracted features $\bm{\Phi}_{c}^{\tilde{X}}$, which encapsulate high-level semantic information from the coarse layer, are incorporated into the dense features ${\bm F}_{d}^{\tilde{X}}$. As illustrated in Fig.\hspace{4pt}\ref{TreeAttention}(b), the incorporated features ${{\bm{\Psi}}}_{d}^{\tilde{X}i}$ for the $i$th dense point are obtained as follows:
\begin{equation} \small
\setlength{\abovedisplayskip}{5pt}
\setlength{\belowdisplayskip}{5pt}
\begin{split}
            {\bm{\Psi}}_{d}^{\tilde{X}i}  = {\bm F}_{d}^{\tilde{X}i} + \bm{\Phi}_{c}^{\tilde{X}j}, j = {\bm \rho}_{d \rightarrow c}^{\tilde{X} i}.
\end{split}
\end{equation}

{The features $\bm{\Psi}_{d}^{\tilde{Y}}$ are obtained via a similar procedure. Subsequently, the attention map $\mathbb{M}_c^{\tilde{X}}$ is utilized to specify the attended regions. As shown in Fig.\hspace{4pt}\ref{TreeAttention}(c), the queries and the corresponding attended regions are highlighted in the same color. For the $i$th query in the coarse layer, the $\mathcal{S}$ key points with the highest attention scores are selected, and their indices are denoted by $\bm{J}^{\tilde{X}i}$. Then, the child points of the $\mathcal{S}$ selected coarse keys constitute the attended regions of the query child points in the dense layer, forming the key points $\bm{K}_{d}^{\tilde{Y}} \in \mathbb{R}^{N_{d}^{\tilde{X}} \times \mathcal{K}_{d}^{\tilde{Y}} \times D}$ within the attended regions:}
\begin{equation} \small
\setlength{\abovedisplayskip}{5pt}
\setlength{\belowdisplayskip}{5pt}
\begin{split}
        \bm{\mathcal{I}} &= \{({\bm \rho}_{c \rightarrow d}^{\tilde{X}i},{\bm \rho}_{c \rightarrow d}^{\tilde{Y}j})|i \in [1,...,N_c^{\tilde{X}}], j \in \bm{J}^{\tilde{X}i}\}, \\
    \bm{K}_{d}^{\tilde{Y}} &= [\bm{K}_{d}^{1}, ... , \bm{K}_{d}^{{|\mathcal{I}|}}],  \bm{K}_{d}^{i} = \bm{\Psi}_{d}^{\tilde{Y}_j},  (i,j) \in \bm{\mathcal{I}}, \\
\end{split}
\end{equation}
where the first and second elements of $\bm{\mathcal{I}}$ represent the child point indices of the coarse queries and child point indices of the $\mathcal{S}$ selected coarse keys, respectively. $\bm{K}_{d}^{i}$ represents the key points within the attended regions of the $i$th dense query point $\bm{\Psi}_{d}^{\tilde{X}i}$.
Such that the child points of the coarse queries only attend to the child points of the selected $\mathcal{S}$ coarse keys.
Then, PTA performs attention to assemble the information of the key points located within the attended regions:
\begin{equation} \small
\setlength{\abovedisplayskip}{5pt}
\setlength{\belowdisplayskip}{5pt}
\begin{split}
    \bm{\Phi}_{d}^{\tilde{X}}, \mathbb{M}_{d}^{\tilde{X}} = {\rm  MA}(\bm{\Psi}_{d}^{\tilde{X}},\bm{K}_{d}^{\tilde{Y}}),
                    \end{split}
\end{equation}
where ${\rm  MA}$ is the attention operation described in Eq. \ref{attention} and $\bm{\Phi}_{d}^{\tilde{X}}$ denotes the extracted features. The attention map $\mathbb{M}_{d}^{\tilde{X}}\! \in \! \mathbb{R}^{N_{d}^{\tilde{X}} \times \mathcal{K}_{d}^{\tilde{Y}}}$ is generated by averaging the attention maps across all heads. By utilizing the formulated key points $\bm{K}_{d}^{\tilde{Y}}$, PTA establishes associations only within the specified areas, allowing it to focus on critical structures. In summary, an integral implementation of PTA is presented in Algorithm \ref{PTA_algo}. 

Similarly, the procedure for obtaining the features $\bm{\Phi}_{d}^{\tilde{Y}} \in \mathbb{R}^{N_{d}^{\tilde{Y}} \times D}$ and the averaged attention map $\mathbb{M}_{d}^{\tilde{Y}} \in \mathbb{R}^{N_{d}^{\tilde{Y}} \times \mathcal{K}_{d}^{\tilde{X}}}$ is performed in parallel.
Finally, the output values $\bm{\Phi}_{l}^{\tilde{X}}$ and $\bm{\Phi}_{l}^{\tilde{Y}}$ from all layers are obtained. The features $\bm{\Phi}_{L_{\tau}}^{\tilde{X}}$ and $\bm{\Phi}_{L_{\tau}}^{\tilde{Y}}$ from the densest layer serve as the final outputs of PTA. 

By utilizing a cross-attention formulation based on PTA, self-attention can be explicitly defined. Considering  $\tilde{{\bm X}}$ as a representative example, PTA  starts with the global attention operation ${\rm MA}({\bm F}_{1}^{\tilde{X}},{\bm F}_{1}^{\tilde{X}})$ in the first layer. {In the subsequent layers, PTA incorporates the spatially coarse features $\bm{\Phi}_{c}^{\tilde{X}}$ to obtain $\bm{\Psi}_{d}^{\tilde{X}}$ and specifies the attended regions according to the attention maps $\mathbb{M}_c^{\tilde{X}}$ in the previous layer. Then, PTA generates the formulated key points $\bm{K}_{d}^{\tilde{X}}$ within the specified attended regions and performs the attention ${\rm MA}(\bm{\Psi}_{d}^{\tilde{X}},\bm{K}_{d}^{\tilde{X}})$ to obtain dense features $\bm{\Phi}_{d}^{\tilde{X}}$ and attention maps $\mathbb{M}_d^{\tilde{X}}$.}

\textcolor{black}{The dynamic attention sparsity of PTA enables each query to be evaluated using only highly relevant key points, thus reducing the computational complexity. 
Without loss of generality, consider two point clouds $\tilde{{\bm X}}  \in \mathbb{R}^{M^{'} \times 3}, \, \tilde{{\bm Y}}  \in \mathbb{R}^{N^{'} \times 3}$. The $L_{\tau}$-layer tree is established such that $N_1^{X} N_1^Y$ is bounded by a constant, irrespective of the number of points in the clouds. At the $l$th layer, the point cloud comprises $N_l$ points, with the maximum $N_l/N_{l+1}$ ratio denoted as $P$, where $P<1$. Notably, $V$ and $P$ remain unrestricted. Following this tree structure, PTA refines attended regions to the child points of $\mathcal{S}$ coarse points and computes attention within these specified regions. The flops of computing PTA are defined as follows:}

\begin{equation} \small
\setlength{\abovedisplayskip}{3pt}
\setlength{\belowdisplayskip}{3pt}
\begin{split}
    case \, 1:& \, L_{\tau} = 2, \\
    Flops &= N_1^X N_1^Y D + N_2^X \mathcal{K}_2 D \\
          &\leqslant N_1^X N_1^Y D + N_2^X \mathcal{S}V D \\
          &= N_1^X N_1^Y D + M^{'} \mathcal{K}_{max} D \quad (\mathcal{K}_{max} = \mathcal{S}V),\\
    case \, 2:& \, L_{\tau} > 2, \\
    Flops &= N_1^X N_1^Y D + \sum_{l=2}^{L_{\tau}} N_l^X \mathcal{K}_l D \\
          &\leqslant N_1^X N_1^Y D + \sum_{l=2}^{L_{\tau}\!-\!1} N_l^X\mathcal{S}\mathbb{N} D + N_{L_{\tau}}^X \mathcal{S}V D \\
          &\leqslant N_1^X N_1^Y D + \sum_{l=2}^{L_{\tau}} N_l^X \mathcal{K}_{max} D \quad
    (\mathcal{K}_{max} = {\rm max}(\mathcal{S}V,\mathcal{S}\mathbb{N})) \\
          &\leqslant N_1^X N_1^Y D + \sum_{l=2}^{L_{\tau}} P^{L_{\tau}-l}M^{'} \mathcal{K}_{max} D \\
          &= N_1^X N_1^Y D + \alpha M^{'} \mathcal{K}_{max} D \quad ( \alpha = \frac{1-P^{L_{\tau}-1}}{1-P}), \\
\end{split}
\end{equation}
\textcolor{black}{where $N_1^X N_1^Y$ is less than a constant independent of the number of points in the point cloud. The computational complexity of PTA is $\mathcal{O}(N \mathcal{K}_{max}D)$. Since $\mathcal{K}_{max}$ is equal to a predefined value ${\rm max}(\mathcal{S}V,\mathcal{S}\mathbb{N})$, the complexity of PTA is linear for the number of points. The inference time and memory usage are intuitively exhibited in the Sec. \ref{efficiency}. }

\textcolor{black}{In general, the dynamic attention sparsity of PTA promotes the attention mechanism to focus on important local structures, facilitating the extraction of local features. Furthermore, PTA hierarchically incorporates coarse features to facilitate multi-scale information aggregation. Through iterative self- and cross-attention, the tree transformer encoder updates the features with contextual information, extracting the conditioned features $\bm{\mathcal{{F}}}^{\tilde{X}}$ and $ \bm{\mathcal{{F}}}^{\tilde{Y}}$ in linear computational complexity.}

\begin{figure*}[t]
    \begin{center}
    \includegraphics[width=18.0cm]{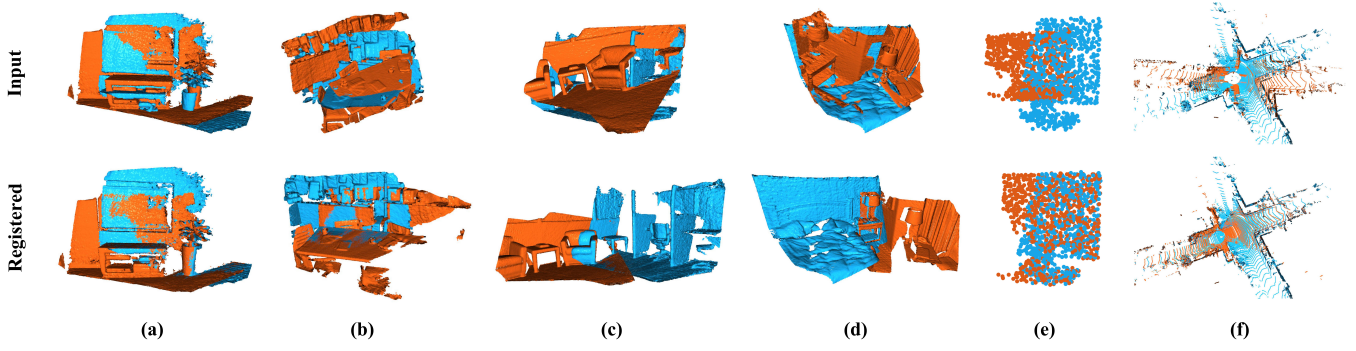}
    \end{center}
       \caption{Qualitative registration results obtained on (a, b) 3DMatch, (c, d) 3DLoMatch, (e) ModelNet40, and (f) KITTI.}
    \label{3DMatch_results}
\end{figure*}

\begin{algorithm}[t] 
    \caption{Point Tree Attention}
    \setstretch{1.1}
    \label{PTA_algo}  
    \begin{spacing}{1.2}
    \begin{algorithmic} 
        \small
      \REQUIRE Tree structures ${\bm \tau}^{\tilde{X}} = \{{\bm \rho}_{c \rightarrow d}^{\tilde{X}}, {\bm \rho}_{d \rightarrow c}^{\tilde{X}}, {\bm C}_{c}^{\tilde{X}}\}$, and ${\bm \tau}^{\tilde{Y}} = \{{\bm \rho}_{c \rightarrow d}^{\tilde{Y}}, {\bm \rho}_{d \rightarrow c}^{\tilde{Y}}, {\bm C}_{c}^{\tilde{Y}}\}$ for point clouds $\tilde{X}$ and $\tilde{Y}$; Input features ${\bm F}^{\tilde{X} }, \, {\bm F}^{\tilde{Y} }$
      \ENSURE  
        The extracted features $\bm{\Phi}_{L_{\tau}}^{\tilde{X}}$ and $\bm{\Phi}_{L_{\tau}}^{\tilde{Y}}$.
        \STATE \textbf{Function} Tree\_layer\_attn($\mathbb{M}_{c}^{s}$, $\bm{\Psi}_{d}^{s}$, $\bm{\Psi}_{d}^{t}$, ${\bm \rho}_{c \rightarrow d}^{s}$, ${\bm \rho}_{c \rightarrow d}^{t}$)
        \STATE \textcolor[RGB]{0,129,0}{\# the attention function in the tree layer}
        \STATE \quad $\bm{\mathcal{I}} \leftarrow specify\_attn\_region(\mathbb{M}_{c}^{s}, {\bm \rho}_{c \rightarrow d}^{s}, {\bm \rho}_{c \rightarrow d}^{t}) $
        \STATE \quad $\bm{K}_{d}^{t} \leftarrow \bm{\Psi}_{d}^{t}[\bm{\mathcal{I}}]$
        \STATE \quad $\bm{\Phi}_{d}^{s}, \mathbb{M}_{d}^{s} \leftarrow {\rm  MA}(\bm{\Psi}_{d}^{s},\bm{K}_{d}^{t})$
        \STATE \quad \textbf{return} $\quad \bm{\Phi}_{d}^{s}, \mathbb{M}_{d}^{s}$
        \STATE \textbf{EndFunction}
        \STATE
        \STATE ${\bm F}_{l}^{\tilde{X}} \leftarrow Feature\_pooling({\bm F}^{\tilde{X}}, {\bm \rho}_{c \rightarrow d}^{\tilde{X}}, {\bm C}_{c}^{\tilde{X}})$ 
        \STATE ${\bm F}_{l}^{\tilde{Y}} \leftarrow Feature\_pooling({\bm F}^{\tilde{Y}}, {\bm \rho}_{c \rightarrow d}^{\tilde{Y}}, {\bm C}_{c}^{\tilde{Y}})$
        \STATE \textcolor[RGB]{0,129,0}{\# conduct global attention in the coarsest layer}
        \STATE $\bm{\Phi}_{1}^{\tilde{X}}, \mathbb{M}_{1}^{\tilde{X}} \leftarrow {\rm  MA}(\bm{F}_{1}^{\tilde{X}},\bm{F}_{1}^{\tilde{Y}})$
        \STATE $\bm{\Phi}_{1}^{\tilde{Y}}, \mathbb{M}_{1}^{\tilde{Y}} \leftarrow {\rm  MA}(\bm{F}_{1}^{\tilde{Y}},\bm{F}_{1}^{\tilde{X}})$
        \STATE \textbf{for} $d = 2$ \textbf{to} $L_{\tau}$ \textbf{do}
        \STATE \quad $c \leftarrow d - 1$
        \STATE \quad \textcolor[RGB]{0,129,0}{\# incorporate the coarse features into the dense features}
        \STATE \quad ${\bm{\Psi}}_{d}^{\tilde{X}}  \leftarrow {\bm F}_{d}^{\tilde{X}} + \bm{\Phi}_{c}^{\tilde{X}}[{\bm \rho}_{d \rightarrow c}^{\tilde{X}}]$
        \STATE \quad ${\bm{\Psi}}_{d}^{\tilde{Y}}  \leftarrow {\bm F}_{d}^{\tilde{Y}} + \bm{\Phi}_{c}^{\tilde{Y}}[{\bm \rho}_{d \rightarrow c}^{\tilde{Y}}]$
        \STATE \quad \textcolor[RGB]{0,129,0}{\# conduct attention within the specified regions}
        \STATE \quad $\bm{\Phi}_{d}^{\tilde{X}}, \mathbb{M}_{d}^{\tilde{X}} \leftarrow Tree\_layer\_attn(\mathbb{M}_{c}^{\tilde{X}}$, $\bm{\Psi}_{d}^{\tilde{X}}$, $\bm{\Psi}_{d}^{\tilde{Y}}$, ${\bm \rho}_{c \rightarrow d}^{\tilde{X}}$, ${\bm \rho}_{c \rightarrow d}^{\tilde{Y}})$
        \STATE \quad $\bm{\Phi}_{d}^{\tilde{Y}}, \mathbb{M}_{d}^{\tilde{Y}} \leftarrow Tree\_layer\_attn(\mathbb{M}_{c}^{\tilde{Y}}$, $\bm{\Psi}_{d}^{\tilde{Y}}$, $\bm{\Psi}_{d}^{\tilde{X}}$, ${\bm \rho}_{c \rightarrow d}^{\tilde{Y}}$, ${\bm \rho}_{c \rightarrow d}^{\tilde{X}})$
      
      \STATE \textbf{end}
      \\  
      \RETURN $\bm{\Phi}_{L_{\tau}}^{\tilde{X}}$, $\bm{\Phi}_{L_{\tau}}^{\tilde{Y}}$  
    \end{algorithmic}  
    \end{spacing}
  \end{algorithm} 

\subsection{Decoder}\label{Decoder}
Inspired by \cite{yew2022regtr}, the conditioned features $\bm{\mathcal{{F}}}^{\tilde{X}}$ and $\bm{\mathcal{{F}}}^{\tilde{Y}}$ are employed to generate corresponding point clouds by means of a two-layer
${\rm MLP}$. Specifically, the corresponding point clouds $\hat{{\bm Y}} \in \mathbb{R}^{M^{'} \times 3}$ and $\hat{{\bm X}} \in \mathbb{R}^{N^{'} \times 3}$ of point clouds $\tilde{{\bm X}}$ and $\tilde{{\bm Y}}$, respectively, are predicted as
\begin{equation} \small
\setlength{\abovedisplayskip}{3pt}
\setlength{\belowdisplayskip}{3pt}
\begin{split}
    \hat{{\bm Y}} = {\rm ReLU}(\bm{\mathcal{{F}}}^{\tilde{X}}{\bm W}_1+{\bm b}_1){\bm W}_2+{\bm b}_2, \\
    \hat{{\bm X}} = {\rm ReLU}(\bm{\mathcal{{F}}}^{\tilde{Y}}{\bm W}_1+{\bm b}_1){\bm W}_2+{\bm b}_2,
\end{split}
\end{equation}
where ${\bm W}_1, {\bm W}_2, {\bm b}_1, {\bm b}_2$ are the learnable parameters in the ${\rm MLP}$. Furthermore, to predict the probabilities that points lie in the overlap regions, the overlap scores $\hat{{\bm o}}^{\tilde{X}} \in \mathbb{R}^{M^{'} \times 1} $ and $\hat{{\bm o}}^{\tilde{Y}} \in \mathbb{R}^{N^{'} \times 1} $ are generated by a single fully connected layer (${\rm FC}$) and the sigmoid activation, as follows:
\begin{equation} \small
\setlength{\abovedisplayskip}{3pt}
\setlength{\belowdisplayskip}{3pt}
\begin{split}
    \hat{{\bm o}}^{\tilde{X}} = {\rm Sigmoid}({\rm FC}(\bm{\mathcal{{F}}}^{\tilde{X}})), \, \hat{{\bm o}}^{\tilde{Y}} = {\rm Sigmoid}({\rm FC}(\bm{\mathcal{{F}}}^{\tilde{Y}})).
\end{split}
\end{equation}

\subsection{Loss Functions}
Our method is trained with three loss functions: an overlap loss $\mathcal{L}_o$, a correspondence loss $\mathcal{L}_c$, and a feature loss $\mathcal{L}_f$.
By introducing coefficients $\lambda_{c}$ and $\lambda_{f}$, the final loss function is constructed and formulated as
\begin{equation} \small
\setlength{\abovedisplayskip}{3pt}
\setlength{\belowdisplayskip}{3pt}
\label{loss_function}
        \mathcal{L} = \mathcal{L}_o + \lambda_{c}\mathcal{L}_c + \lambda_{f} \mathcal{L}_f.
\end{equation}

\textbf{Overlap loss}. $\mathcal{L}_o$ measures the consistency between the ground-truth overlap labels ${\bm o}^{\tilde{X}}, \, {\bm o}^{\tilde{Y}}$ and the predicted overlap scores $\hat{{\bm o}}^{\tilde{X}}, \, \hat{{\bm o}}^{\tilde{Y}}$. $\mathcal{L}_o\!=\!\mathcal{L}_o^{X}\!+\!\mathcal{L}_o^{Y}$, and $\mathcal{L}_o^{X}$ is defined as
\begin{equation} \small
\setlength{\abovedisplayskip}{3pt}
\setlength{\belowdisplayskip}{3pt}
    \begin{split}
        \mathcal{L}_o^{X} \!&\!= \!\frac{-1}{M^{'}} \sum_{i=1}^{M^{'}}[{\bm o}^{\tilde{X}_{i}} \!\times \! {\rm log} \hat{{\bm o}}^{\tilde{X}_{i}}  
    \!+\!(1\!-\!{\bm o}^{\tilde{X}_{i}}))\! \times \!{\rm log} (1\!-\!\hat{{\bm o}}^{\tilde{X}_{i}})]. 
    \end{split}
\end{equation}

The overlap labels ${\bm o}^{\tilde{X}}, \, {\bm o}^{\tilde{Y}}$ are obtained by downsampling the overlap labels ${\bm o}^{X}, \, {\bm o}^{Y}$ of point clouds ${\bm X}, \, {\bm Y}$, the overlap labels ${\bm o}^{X_{i}}$ for point $X_{i} \in {\bm X}$ are defined as:
\begin{equation} \small
\setlength{\abovedisplayskip}{3pt}
\setlength{\belowdisplayskip}{3pt}
    \begin{split}
        {\bm o}^{X_{i}} =\left\{\begin{array}{ll}
1, & \left\|{\bm T}_X^Y\left(X_{i}\right)-{\rm NN}\left({\bm T}_X^Y\left(X_{i}\right), {\bm Y}\right)\right\|<r_{o} \\
0, & \text { otherwise }
\end{array}\right.,
    \end{split}
\end{equation}
where ${\bm T}_X^Y$ denotes the transformation from ${\bm X}$ to ${\bm Y}$, ${\rm NN}(\cdot)$ indicates the spatial nearest neighbor, and $r_{o}$ is the predefined overlap threshold.

\textbf{Correspondence loss}. $\mathcal{L}_c$ measures the correctness of the predicted corresponding point clouds in the overlapping regions based on the $\ell^{1}$ loss. $\mathcal{L}_c=\mathcal{L}_c^{X} + \mathcal{L}_c^{Y}$, with the correspondence loss $\mathcal{L}_c^{X}$ defined as
\begin{equation} \small
\setlength{\abovedisplayskip}{3pt}
\setlength{\belowdisplayskip}{3pt}
    \begin{split}
        \mathcal{L}_{c}^{X}=\frac{1}{\sum_{i=1}^{M^{'}} {\bm o}^{\tilde{X}_{i}}} \sum_{i=1}^{M^{'}} {\bm o}^{\tilde{X}_{i}}\left|{\bm T}_X^Y\left(\tilde{X}_{i}\right)-\hat{Y}_{i}\right|,
    \end{split}
\end{equation}
where ${\bm T}_X^Y$ is the ground-truth transformation from ${\bm X}$ to ${\bm Y}$.

\textbf{Feature loss}. $\mathcal{L}_f$ measures the discriminative power of the extracted features based on the InfoNCE loss \cite{oord2018representation}. $\mathcal{L}_f = \mathcal{L}_f^{X} + \mathcal{L}_f^{Y}$, with $\mathcal{L}_f^{X}$ defined as
\begin{equation} \small
\setlength{\abovedisplayskip}{3pt}
\setlength{\belowdisplayskip}{3pt}
    \begin{split}
        \mathcal{L}_{f}^{X}=-\mathbb{E}_{x \in  \mathcal{X}}& \left[\log \frac{f\left({x}, {p}_{x}\right)}{f\left({x}, {p}_{{x}}\right)+\sum_{{n}_{{x}}} f\left({x}, {n}_{{x}}\right)}\right],  \\
        f(x,c) &= {\rm exp}({\bm{\mathcal{{F}}}^{x}}^{T}{W_f}\bm{\mathcal{{F}}}^{c}),
    \end{split}
\end{equation}
where $\mathcal{X}$ denotes the set of points $\mathcal{X} \subseteq \tilde{{\bm X}}$ with a correspondence in $\tilde{{\bm Y}}$; $\bm{\mathcal{{F}}}^{x}$ indicates the extracted features for point $x$. $p_{x}$ and $n_{x}$ denote the positive and negative points in $\tilde{{\bm Y}}$, which are selected based on the positive and negative margins $(r_p, r_n)$; and $W_f$ is a learnable linear transformation.

\section{Experimental Results}

\subsection{Implementation Details}
\textcolor{black}{\textbf{Network architecture}. Due to variations in density and size among point clouds from different benchmarks, we employ slightly different backbones and tree structures in our experiments. Specifically, we utilize a 4-stage backbone for 3DMatch and ModelNet40, while opting for a 5-stage backbone for KITTI due to its significantly larger point clouds. Our PTT incorporates a 6-layer tree transformer encoder, with PTA configured to employ 8 heads. In the PTA module, we employ a 3-layer tree structure for point cloud representation, setting $\mathbb{N}$ and $\bm{V}$ to 4, and $m$. Here, $m$ denotes the voxel distance utilized in the final downsampling layer of the KPConv network, configured as 0.2 for 3DMatch, 0.12 for ModelNet, and 4.8 for KITTI datasets. Leveraging these constructed tree structures, PTA hierarchically specifies attended regions by selecting the child points of the top $\mathcal{S}=8$ coarse points with the highest attention scores.}

\textcolor{black}{\textbf{Training and testing}. The loss functions employ $\lambda_{c}=1$, $\lambda_{f}=0.1$, and $(r_p,r_n)=(m,2m)$. Our PTT is implemented and evaluated using PyTorch \cite{paszke2019pytorch} on hardware consisting of an Intel I7-10700 CPU paired with an RTX 3090 graphics card. The models are trained utilizing AdamW \cite{loshchilov2017decoupled} for 70 epochs on 3DMatch, 400 epochs on ModelNet, and 200 epochs on KITTI, with a weight decay of $1e\!-\!4$. Batch sizes are 1 for 3DMatch/KITTI and 4 for ModelNet. The initial learning rate is $1e\!-\!4$, with halving scheduled every 20 epochs on 3DMatch, every 100 epochs on ModelNet, and every 50 epochs on KITTI. The same data augmentation as in \cite{huang2021predator} is adopted.}

\begin{figure*}[t]
    \begin{center}
    \includegraphics[width=18.0cm]{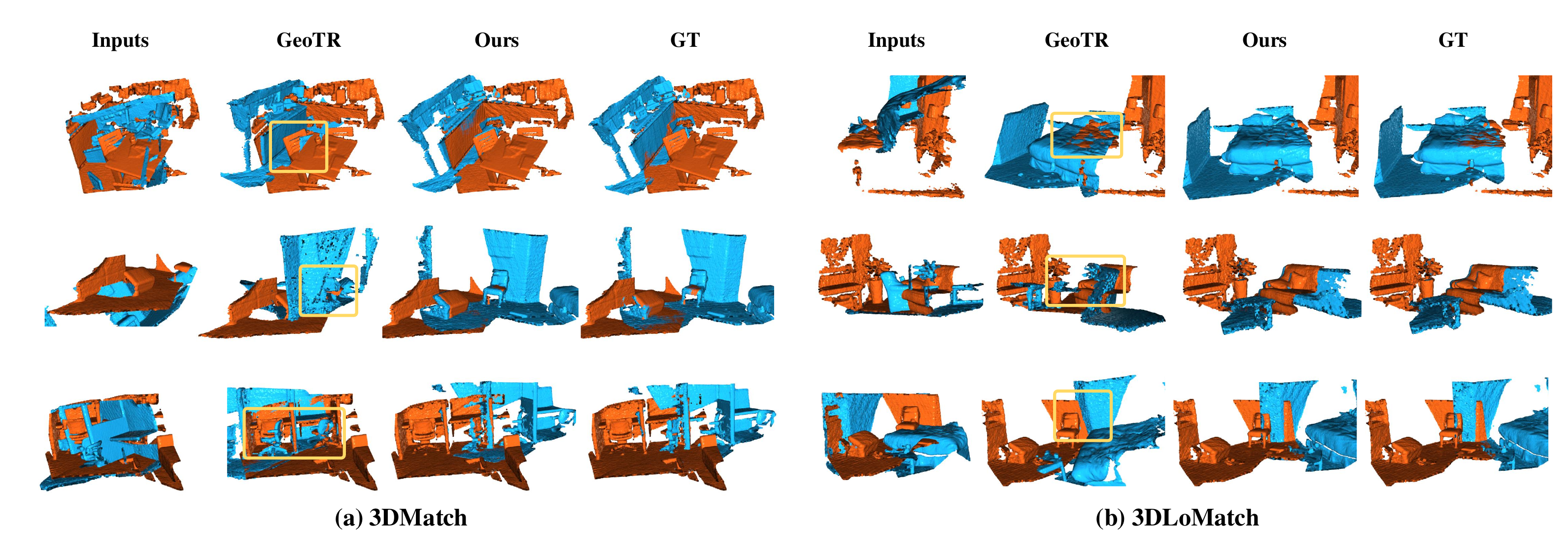}
    \end{center}
    \setlength{\abovecaptionskip}{-0.15cm}
   \caption{Comparison of the registration results on (a) 3DMatch and (b) 3DLoMatch benchmarks.}
    \label{3DMatch_Comp_results}
\end{figure*}

\subsection{Registration Performance on 3DMatch}
\textbf{3DMatch}. To demonstrate the real-world point cloud registration performance of our method, experiments are conducted on 3DMatch \cite{zeng20173dmatch}.  The 3DMatch  dataset is a real-world registration dataset, in which 46 scenes are designed for training, and the remaining 16 scenes are evenly allocated for validation and testing. The comparison methods are evaluated on both the 3DMatch ($> \!30\%$ overlap) \cite{zeng20173dmatch} and 3DLoMatch ($10\!- \!30\%$ overlap) \cite{huang2021predator} benchmarks.

\textbf{Comparison methods}. 
Our method PTT is compared with the latest approaches: RegTR \cite{yew2022regtr}, Lepard \cite{li2022lepard}, SC$^2$PCR \cite{chen2022sc2}, GeoTransformer (GeoTR) \cite{qin2022geometric}, OIF-PCR \cite{yangone}, VBReg \cite{jiang2023robust}, and BUFFER \cite{ao2023buffer}. Furthermore, other comparison methods include
3DSN \cite{gojcic2019perfect}, FCGF \cite{choy2019fully}, CG-SAC \cite{9052691}, D3Feat \cite{bai2020d3feat}, DGR \cite{choy2020deep}, PCAM \cite{cao2021pcam},   DHVR\cite{lee2021deep}, Predator \cite{huang2021predator}, and CoFiNet \cite{yu2021cofinet}.
In the correspondence-based methods based on RANSAC, the number of interest points is set to 5000.

\textbf{Evaluation metrics}. The methods are evaluated through various performance metrics, as per \cite{yew2022regtr}. These include the \textit{relative rotation error}
$\rm RRE$ 
, \textit{relative translation error} $\rm RTE$ 
, and \textit{registration recall} $\rm RR$ (the percentage of successful alignments, where a correspondence with a root-mean-square-error below 0.2 m is considered successful).
Notably, as the PTT directly predicts corresponding coordinates without finding correspondences, the inlier ratio and feature matching recall are not reported.

\begin{table}
                \center
        \footnotesize
        \caption{Performance on the 3DMatch and 3DLoMatch benchmarks. The RRE is given in $^{\circ}$, the RTE in $m$, and the RR in $\%$. The three best results are  highlighted in \textbf{\textcolor{red}{red}}, \textbf{\textcolor{green}{green}}, and  \textbf{\textcolor{blue}{blue}}.}
        \label{3DMatch_table}
        \begin{tabular}{l@{\hspace{3pt}}c@{\hspace{3pt}}|c@{\hspace{4pt}}c@{\hspace{4pt}}c@{\hspace{4pt}}|c@{\hspace{4pt}}c@{\hspace{4pt}}c@{\hspace{4pt}}}
            \toprule[1.5pt]
            \multirow{2}*{\small Method}  & \multirow{2}*{\small Reference}  & \multicolumn{3}{c|}{\small 3DMatch}
            & \multicolumn{3}{c}{\small 3DLoMatch}
            \\
            ~ & ~   &{\scriptsize {$\rm RRE$}} & {\scriptsize ${\rm RTE}$} & {\scriptsize {$\rm RR$}} &{\scriptsize {$\rm RRE$}} &{\scriptsize ${\rm RTE}$} & {\scriptsize {$\rm RR$}}  \\
            \midrule[1pt]
                        3DSN  \cite{gojcic2019perfect}        & CVPR 2019            &2.19            &0.071    &78.4         & 3.52   &0.103 &33.0             \\
            FCGF \cite{choy2019fully} &CVPR 2019             &2.14           &0.070 &85.1          &3.74 &0.100    &40.1          \\
            CG-SAC\cite{9052691} & T-GE 2020  &2.42 &0.076 & 87.5 & 3.86 &0.109 &64.0
            \\
            D3Feat \cite{bai2020d3feat} &CVPR 2020           & 2.16          &   0.067 &81.6       &3.36 &0.103     &37.2               \\
            DGR \cite{choy2020deep} & {CVPR 2020            }& {2.10           }& {0.067            } & 85.3 & {3.95} &0.113  & 48.7        \\
            PCAM \cite{cao2021pcam} & {ICCV 2021            }& 1.80& {\textbf{\textcolor{blue}{0.059}}}& 85.5 &{3.52} &0.099        & 54.9 \\
            DHVR \cite{lee2021deep} & {ICCV 2021            }& {2.25            }& {0.078            }& 91.9 &4.97 &0.123 & 65.4       \\
            Predator  \cite{huang2021predator}         &CVPR 2021           &2.02             &0.064 &89.0            &3.04         &0.093&62.5       \\
            CoFiNet \cite{yu2021cofinet} & {Neurips 2021            }  & {2.44            }& {0.067            }&89.3 &{5.44} &0.155& 67.5         \\
            RegTR \cite{yew2022regtr} & {CVPR 2022           }& {\textbf{\textcolor{green}{1.57}}}&{\textbf{\textcolor{green}{0.049}}}& 92.0& {\textbf{\textcolor{green}{2.83}}} &{\textbf{\textcolor{green}{0.077}}} & 64.8        \\
            Lepard \cite{li2022lepard} & {CVPR 2022            }& {2.48            }& {0.072            }& {\textbf{\textcolor{green}{93.5}}} &4.10 &0.108 & 69.0       \\
            SC$^2$PCR \cite{chen2022sc2} & {CVPR 2022            }& {2.08            }& {0.065            }& {\textbf{\textcolor{blue}{93.3}}} &3.46 &0.096 & 69.5       \\
            GeoTR \cite{qin2022geometric} & {CVPR 2022            }& {\textbf{\textcolor{blue}{1.72}}}& {0.062            }& 92.0 &{\textbf{\textcolor{blue}{2.93}}} &{\textbf{\textcolor{blue}{0.089}}} & {\textbf{\textcolor{green}{75.0}}}       \\
            VBReg \cite{jiang2023robust} & {CVPR 2023            }& 2.04 & 0.065 & {\textbf{\textcolor{green}{93.5}}} &3.48 &0.096 & 69.9       \\
            BUFFER \cite{ao2023buffer} & {CVPR 2023            }& 1.85 & {\textbf{\textcolor{blue}{0.059}}} & 93.2 &3.09 &0.101 & {\textbf{\textcolor{blue}{71.8}}}       \\
            Ours & {-            }& {\textbf{\textcolor{red}{1.49}}}& {\textbf{\textcolor{red}{0.043}}}&{\textbf{\textcolor{red}{95.4}}} &  {\textbf{\textcolor{red}{2.26}}} &
            {\textbf{\textcolor{red}{0.067}}} &{\textbf{\textcolor{red}{76.3}}}\\
            \bottomrule[1.5pt]
        \end{tabular}
    \end{table}

The qualitative results and comparisons are presented in Fig. \ref{3DMatch_results}(a-d) and Fig. \ref{3DMatch_Comp_results}, while the quantitative comparisons are summarized in Table \ref{3DMatch_table}. The results show that our method precisely aligns a pair of real-world point clouds even at low overlap rates and outperforms the other methods on both 3DMatch and 3DLoMatch. 
{Specifically, the $\rm RR$ of our method is better than those of Lepard and SC$^2$PCR by 7.3\% and 6.8\%, respectively, on the 3DLoMatch benchmark. Additionally, the PTT outperforms GeoTR in ${\rm RR}$ by 3.5\% and 1.3\% on the 3DMatch and 3DLoMatch benchmarks, respectively, while also reducing the RRE and RTE. Even compared with recent VBReg and BUFFER, the PTT still achieves superior performance. 
These results demonstrate that the learned attended regions and the guidance of the coarse features facilitate local feature extraction, enabling our method to precisely align real-world point clouds.}

\begin{table}
                \center
        \footnotesize
        \caption{Performance on the ModelNet40 benchmark. The RRE is given in $^{\circ}$. The three best results are  highlighted in \textbf{\textcolor{red}{red}}, \textbf{\textcolor{green}{green}}, and \textbf{\textcolor{blue}{blue}}.}
        \label{ModelNet_table}
        \begin{tabular}{l@{\hspace{3pt}}c@{\hspace{3pt}}|c@{\hspace{3pt}}c@{\hspace{3pt}}c@{\hspace{3pt}}|c@{\hspace{3pt}}c@{\hspace{3pt}}c@{\hspace{3pt}}}
            \toprule[1.5pt]
            \multirow{2}*{\small Method}  & \multirow{2}*{\small Reference}  & \multicolumn{3}{c|}{\small ModelNet}
            & \multicolumn{3}{c}{\small ModelLoNet}
            \\
            ~ & ~  & {\scriptsize {$\rm RRE $}} &{\scriptsize {$\rm RTE$}} & {\scriptsize {$\rm CD$}}& {\scriptsize {$\rm RRE $}} &{\scriptsize {$\rm RTE$}} & {\scriptsize {$\rm CD$}}   \\
            \midrule[1pt]
            ICP \cite{besl1992method} & SPIE 1992 & 27.2  & 0.280 & 0.0230 &47.5&0.479&0.0521 \\
            FGR \cite{zhou2016fast} & ECCV 2016  &  30.8  &  0.192 &  0.0241 &58.7&0.557&0.0517 \\
            PNetLK \cite{aoki2019pointnetlk} & {CVPR 2019            }& 29.7& 0.297 &  0.0235& 48.5&0.507&0.0367      \\
            DCP-v2 \cite{wang2019deep} & {ICCV 2019            }&11.9& {0.171            }& 0.0117&16.5&0.300&0.0268    \\
            RPM-Net \cite{yew2020rpm} & {CVPR 2020} &1.71 &0.018& {\textbf{\textcolor{blue}{8.5e-4}}}&7.34&0.124&{\textbf{\textcolor{blue}{0.0050}}} \\
            Predator \cite{huang2021predator} & CVPR 2021          &1.73   & 0.019    &8.9e-4 &5.24&0.132&0.0083                 \\
            RegTR \cite{yew2022regtr} & {CVPR 2022           }& {\textbf{\textcolor{blue}{1.47}}}& {\textbf{\textcolor{blue}{0.014}}}&{\textbf{\textcolor{green}{7.8e-4}}}  &{\textbf{\textcolor{blue}{3.93}}}&{\textbf{\textcolor{blue}{0.087}}}&{\textbf{\textcolor{green}{0.0037}}}       \\
            UDPReg \cite{mei2023unsupervised} & CVPR 2023 & {\textbf{\textcolor{green}{1.33}}} & {\textbf{\textcolor{green}{0.011}}} & 0.0306 & {\textbf{\textcolor{green}{3.58}}} & {\textbf{\textcolor{green}{0.069}}} & 0.0416 \\
            Ours & {-            }&{\textbf{\textcolor{red}{1.30}}}& {\textbf{\textcolor{red}{0.010}}} &{\textbf{\textcolor{red}{7.5e-4}}}&{\textbf{\textcolor{red}{3.65}}}&{\textbf{\textcolor{red}{0.068}}}&{\textbf{\textcolor{red}{0.0031}}}\\
            \bottomrule[1.5pt]
        \end{tabular}
            \end{table}

\begin{figure}[t]
    \begin{center}
    \includegraphics[width=8.2cm]{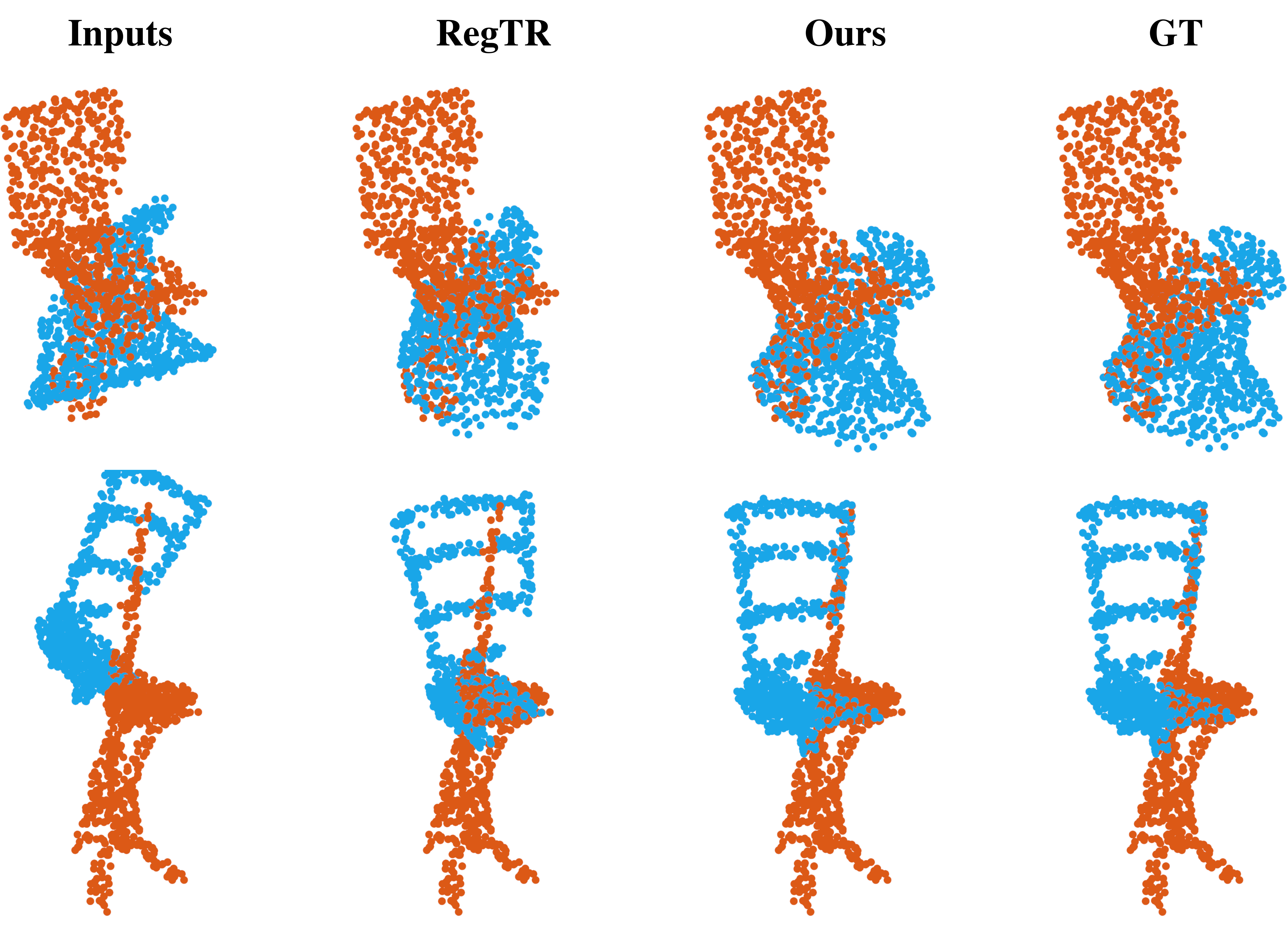}
    \end{center}
    \setlength{\abovecaptionskip}{-0.15cm}
   \caption{Comparison of the registration results on ModelLoNet benchmark.}
    \label{ModelNet_Comp_results}
\end{figure}

\subsection{Registration Performance on ModelNet40} \label{ModelNet40 ex}

\textbf{ModelNet40}. 
\textcolor{black}{ModelNet40 dataset \cite{wu20153d} includes 12,311 CAD meshed models in 40 categories, of which 5,112 samples are used for training, 1,202 samples are used for validation, and 1,266 samples are used for testing. Utilizing processed data from \cite{qi2017pointnet}, which uniformly samples 2048 points on each CAD model's surface, we normalize the CAD model into a unit sphere. Subsequently, source and target point clouds are generated as follows: Partial scans are produced according to \cite{yew2020rpm}. The source point cloud is then randomly transformed with a rotation within $[0, 45^{\circ}]$ and a translation within $[-0.5,0.5]$. Both point clouds are then jittered with a noise sampled from $N(0,0.01)$ and clipped to $[-0.05,0.05]$. At last, $717$ points are randomly sampled from each point cloud as the final point cloud pair. The compared methods are evaluated under two partial overlap settings: ModelNet, which has 73.5\% pairwise overlap on average, and ModelLoNet, which possesses a 53.6\% average overlap rate.}

\textbf{Comparison methods}. 
The PTT is compared with the latest approach UDPReg \cite{mei2023unsupervised}, RegTR \cite{yew2022regtr}; the comparison methods also include ICP \cite{besl1992method}, FGR \cite{zhou2016fast}, PointNetLK (PNetLK) \cite{aoki2019pointnetlk}, DCP-v2 \cite{wang2019deep}, IDAM \cite{li2020iterative}, RPM-Net \cite{yew2020rpm}, and Predator \cite{huang2021predator}. Specifically, Predator samples 450 points.

\textbf{Evaluation metrics}. The performance of the comparison methods is evaluated in terms of the $\rm RRE$, the $\rm RTE$, and the \textit{Chamfer distance} $\rm CD$ between the registered scans.

\begin{table}
                \center
        \small
        \caption{Performance on the KITTI benchmark.The three best results are  highlighted in \textbf{\textcolor{red}{red}}, \textbf{\textcolor{green}{green}}, and \textbf{\textcolor{blue}{blue}}.}
        \label{KITTI_table}
        \begin{tabular}{l@{\hspace{10pt}}c@{\hspace{10pt}}|c@{\hspace{4pt}}c@{\hspace{4pt}}c@{\hspace{4pt}}}
            \toprule[1.5pt]
            {\small Method}  & {\small Reference}  & 
 {\small {$\rm RRE (^{\circ})$}} &{\small {$\rm RTE (m)$}} & {\small ${\rm RR (\%)}$}\\
            \midrule[1pt]
                        FCGF \cite{choy2019fully} & {ICCV 2019           } & 0.30  &0.095& 96.6 \\
            FMR \cite{huang2020feature} & {CVPR 2020            } & 1.49 &0.660& 90.6 \\
            DGR \cite{choy2020deep} & {CVPR 2020            } & 0.37&0.320 & 98.7 \\
            D3Feat \cite{bai2020d3feat} & {CVPR 2020            } & 0.30 &0.072 & {\textbf{\textcolor{red}{99.8}}} \\
            HRegNet \cite{lu2021hregnet} & {ICCV 2021            } & 0.29 &0.120 & 99.7 \\
            SpinNet \cite{ao2021spinnet} & {CVPR 2021            }& 0.47&0.099  & 99.1 \\
            Predator \cite{huang2021predator} & {CVPR 2021            } & {\textbf{\textcolor{blue}{0.28}}} &{\textbf{\textcolor{blue}{0.068}}}& {\textbf{\textcolor{red}{99.8}}} \\
            CoFiNet \cite{yu2021cofinet} & {Neurips 2021            } & 0.41 &0.082& {\textbf{\textcolor{red}{99.8}}} \\
            SC$^2$PCR \cite{chen2022sc2} & CVPR 2022 & 0.32 & 0.072 & 99.6 \\
            GeoTR \cite{qin2022geometric} & CVPR 2022 & {\textbf{\textcolor{green}{0.24}}} & {\textbf{\textcolor{blue}{0.068}}} & {\textbf{\textcolor{red}{99.8}}} \\
            OIF-PCR \cite{yangone} & Neurips 2022 & {\textbf{\textcolor{red}{0.23}}} & {\textbf{\textcolor{green}{0.065}}} & {\textbf{\textcolor{red}{99.8}}} \\
            MAC \cite{zhang20233d} & CVPR 2023 & 0.40 & 0.084 & 99.5 \\
            RegFormer \cite{liu2023regformer} & ICCV 2023 & {\textbf{\textcolor{green}{0.24}}} & 0.084 & {\textbf{\textcolor{red}{99.8}}} \\
            Ours & {-            }&{\textbf{\textcolor{red}{0.23}}}& {\textbf{\textcolor{red}{0.063}}}& {\textbf{\textcolor{red}{99.8}}}\\
            \bottomrule[1.5pt]
        \end{tabular}
    \end{table}

\begin{figure}[t]
    \begin{center}
    \includegraphics[width=8.2cm]{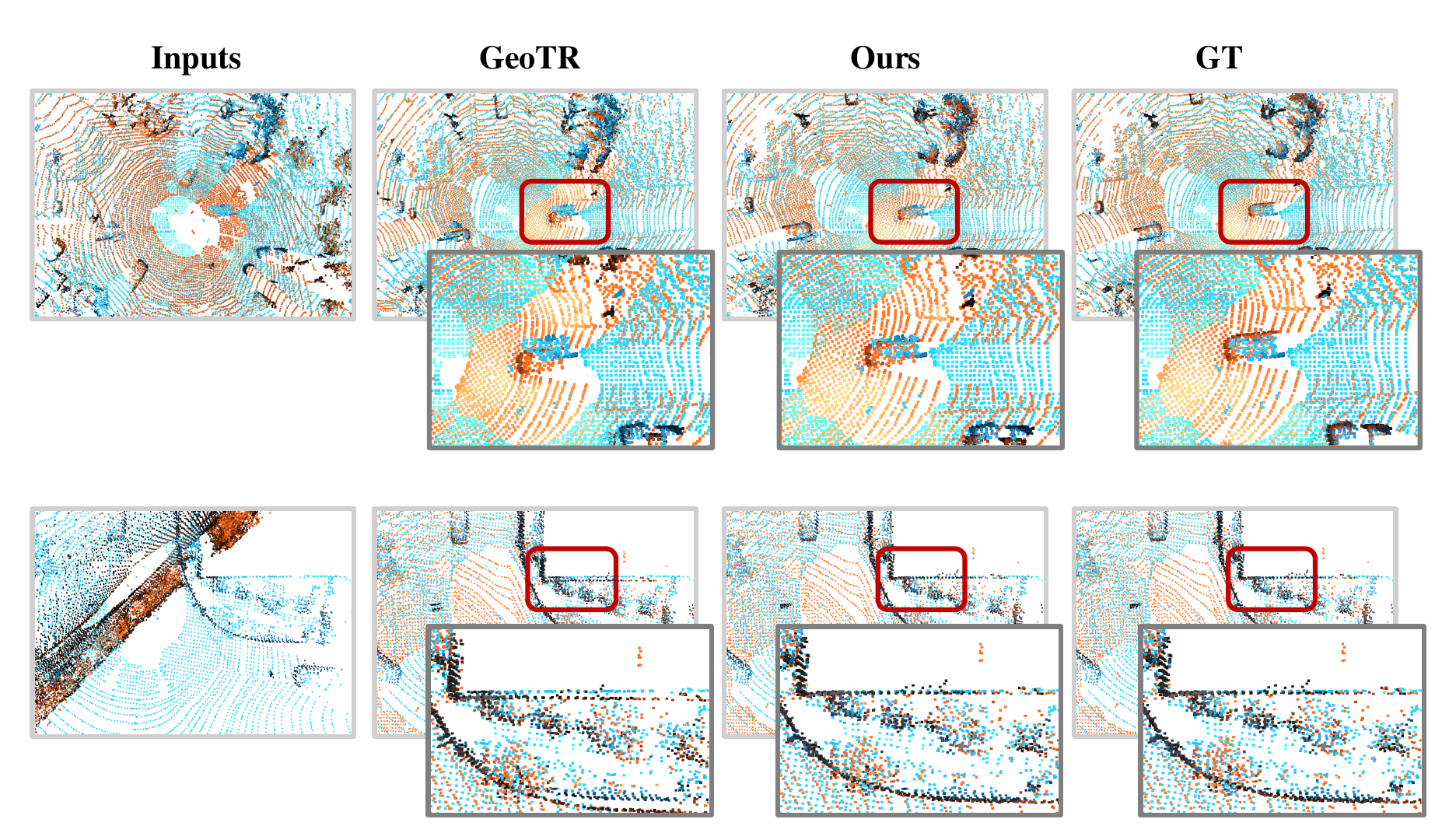}
    \end{center}
    \setlength{\abovecaptionskip}{-0.15cm}
   \caption{Comparison of the registration results on KITTI benchmark.}
    \label{KITTI_Comp_results}
\end{figure}

The qualitative results and comparisons are shown in Fig.\hspace{4pt}\ref{3DMatch_results}(e) and Fig. \ref{ModelNet_Comp_results}, and the quantitative comparisons are summarized in Table \ref{ModelNet_table}. The results demonstrate that our method achieves precise registration of partially visible point clouds and outperforms other methods in terms of all metrics.
Despite the utilization of surface normal information
in RPM-Net \cite{yew2020rpm}, our method
outperforms it in accuracy by 20--50\% on both the ModelNet and ModelLoNet benchmarks. Even in comparison to RegTR and UDPReg, the PTT still achieves superior performance.
These experimental results verify that the proposed PTT has an enhanced capability to capture important local structures. Additionally, {the feature extraction guidance provided by the coarse features } prevents our method from excessively focusing on subtle characteristics that are sensitive to noise.

\subsection{Registration Performance on KITTI} \label{KITTI ex}

\textbf{KITTI}. To demonstrate the performance of our method on a large-scale point cloud dataset, the PTT and the baseline methods are evaluated on the KITTI \cite{geiger2012we} dataset. 
The KITTI dataset contains 11 sequences of LiDAR-scanned outdoor driving scenarios, of which scenarios 0--5 are used for training, 6--7 are used for validation, and 8--10 are used for testing.
Following \cite{huang2021predator}, only point cloud pairs that are at most $10 m$ away from each other are utilized for evaluation.

\textbf{Comparison methods}. 
Our method PTT is compared with the latest approaches MAC \cite{zhang20233d}, RegFormer \cite{liu2023regformer}, SC$^2$PCR \cite{chen2022sc2}, GeoTransformer (GeoTR) \cite{qin2022geometric}, and OIF-PCR \cite{yangone}; the baseline methods also include FCGF \cite{choy2019fully}, FMR \cite{huang2020feature}, DGR \cite{choy2020deep}, D3Feat \cite{bai2020d3feat}, HRegNet \cite{lu2021hregnet}, SpinNet \cite{ao2021spinnet}, Predator \cite{huang2021predator}, and CoFiNet \cite{yu2021cofinet}.

\textbf{Evaluation metrics}. Following \cite{yew2022regtr}, the performance of each method is evaluated using the
$\rm RRE$, $\rm RTE$, and $\rm RR$ (the percentage of successful alignments, whose $\rm RRE$ and $\rm RTE$ values are below $5^{\circ}$ and $2m$, respectively).

The qualitative results and comparisons are shown in Fig.\hspace{4pt}\ref{3DMatch_results}(f) and Fig. \ref{KITTI_Comp_results}, and the quantitative comparisons are summarized in Table \ref{KITTI_table}. The results indicate that our method achieves the highest accuracy on the KITTI benchmark. These experimental results demonstrate that our method possesses an enhanced ability to capture important local features, thereby achieving superior performance.

\subsection{Efficiency Evaluation} \label{efficiency}
\setlength{\tabcolsep}{0.4mm}
{
    \begin{table*}[t!]
    \setlength{\abovecaptionskip}{4pt}
        \center
        \caption{Computational time in seconds on the 3DMatch benchmark.}
        \label{time}
        \begin{tabular}{c@{\hspace{3.5pt}}c@{\hspace{3.5pt}}c@{\hspace{3.5pt}}c@{\hspace{3.5pt}}c@{\hspace{3.5pt}}c@{\hspace{3.5pt}}c@{\hspace{3.5pt}}c@{\hspace{3.5pt}}c@{\hspace{3.5pt}}c@{\hspace{3.5pt}}c@{\hspace{3.5pt}}c@{\hspace{3.5pt}}c@{\hspace{3.5pt}}c@{\hspace{3.5pt}}c@{\hspace{3.5pt}}c@{\hspace{3.5pt}}}
            \toprule[1.5pt]
            3DSN & FCGF & CG-SAC & D3Feat & DGR & PCAM & DHVR & Predator & CoFiNet&RegTR & Lepard & SC$^2 $PCR & GeoTR & VBReg & BUFFER & Ours
            \\
                        \midrule[0.5pt]
            30.234        &1.562       & 0.263     & 0.916 & 1.741 &1.786 & 3.43 &1.572   &1.134 &0.103  &0.522 &0.380 & 1.523 & 0.223 &0.196 & 0.123          \\
            \bottomrule[1.5pt]
        \end{tabular}
    \end{table*}
}

\begin{figure*}[t!]
    \begin{center}
    \includegraphics[width=18cm]{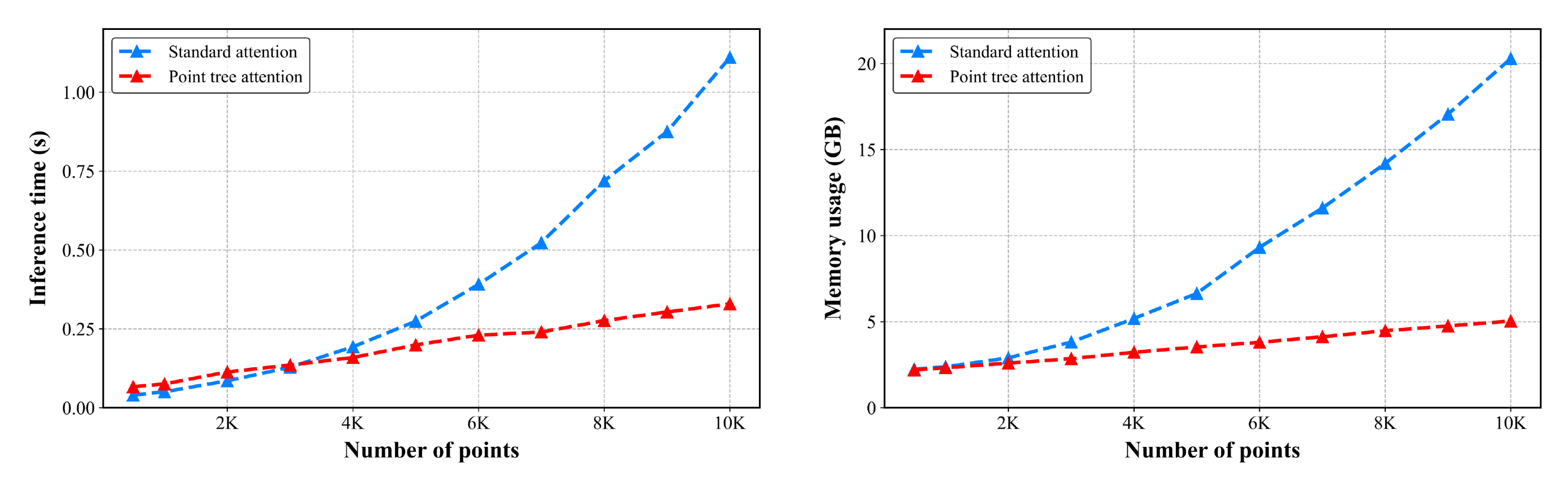}
    \end{center}
    \setlength{\abovecaptionskip}{-0.35cm}
   \caption{Comparison of the inference time and memory usage between standard attention \cite{vaswani2017attention} and our Point tree attention. As the number of points increases, both the inference time and memory usage of standard attention grow quadratically, whereas those of PTA increase linearly.}
    \label{time_memory}
\end{figure*}

\textcolor{black}{\textbf{Comparison with the registration methods}. The inference time of the comparison methods is evaluated using the 3DMatch benchmark. Experiments are conducted on a desktop computer with an Intel I7-10700 CPU and an Nvidia RTX 3090 GPU. The comparison methods include 
3DSN \cite{gojcic2019perfect}, FCGF \cite{choy2019fully}, CG-SAC \cite{9052691}, D3Feat \cite{bai2020d3feat}, DGR \cite{choy2020deep}, PCAM \cite{cao2021pcam},    DHVR\cite{lee2021deep}, Predator \cite{huang2021predator}, CoFiNet \cite{yu2021cofinet}, RegTR \cite{yew2022regtr}, Lepard \cite{li2022lepard}, SC$^2$PCR \cite{chen2022sc2}, GeoTR \cite{qin2022geometric}, VBReg \cite{jiang2023robust}, and BUFFER \cite{ao2023buffer}. As illustrated in Table \ref{time}, PTT is positioned as the second fastest method, achieving pipeline completion in less than 130 ms, a timeframe conducive to many applications. Moreover, it lags merely 20ms behind the fastest method, RegTR, while notably outperforming RegTR in accuracy. Additionally, compared to recent methods, GeoTR, VBReg, and BUFFER, our approach demonstrates superior performance in both efficiency and accuracy. 
It is noteworthy that our inference time is achieved solely using a basic CUDA kernel without extensive optimizations, while the standard attention mechanism is equipped with well-optimized dense GPU matrix operations. }

\textcolor{black}{\textbf{Comparison with the standard attention}. The illustration in Fig. \ref{time_memory} showcases the inference time and memory usage of the PTT network respectively employing the standard attention and the PTA mechanisms. As the number of points increases, the inference time and memory consumption of our PTA exhibit linear growth, contrasting with the quadratic increase observed in standard attention. Notably, PTA generally demonstrates lower inference times compared to standard attention, alongside consistently reduced memory utilization. At a point count of 10K, standard attention necessitates over 1.1s for inference time and 20GB for memory usage. In comparison, the network with the PTA can efficiently align point clouds, demanding less than 0.36s for inference time and a mere 5GB of memory usage. This signifies a noteworthy reduction of 67\%–75\% in both inference time and memory usage when contrasted with standard attention. Furthermore, PTA is currently implemented employing a basic CUDA kernel without additional CUDA optimizations, and it exhibits proficiency in extracting intricate local features and enhancing registration accuracy. Overall, PTA demonstrates superior accuracy and computational efficiency compared to standard attention.
}

\begin{figure*}[ht]
    \begin{center}
    \includegraphics[width=18.2cm]{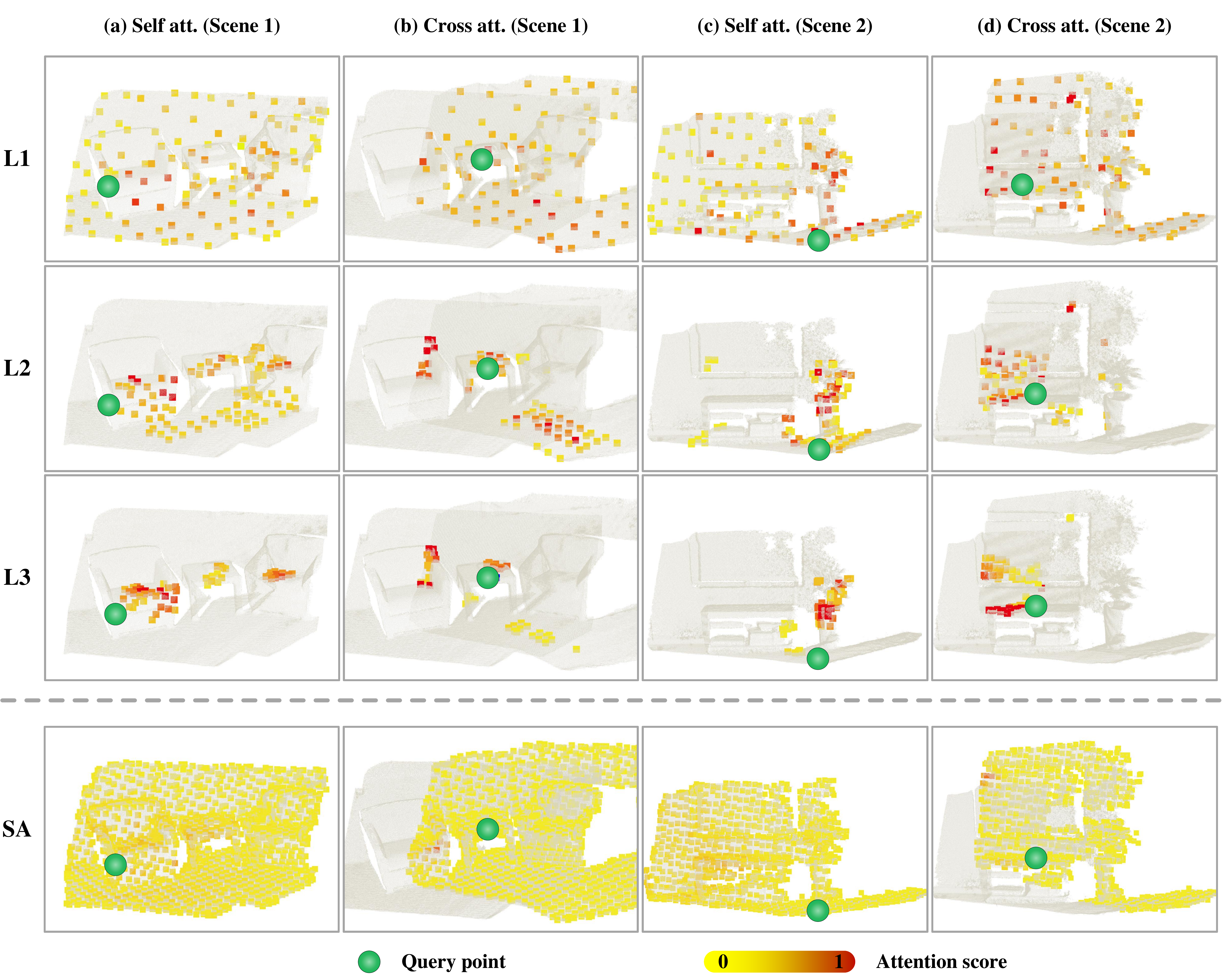}
    \end{center}
    \setlength{\abovecaptionskip}{-0.15cm}
   \caption{Visualization of attention weights on the 3DMatch dataset. L1, L2, and L3 correspondingly represent the first, second, and third layers of the PTA, while SA denotes the standard attention mechanism.}
    \label{attention}
\end{figure*}

\subsection{Attention Visualizations}
\textcolor{black}{To comprehensively demonstrate the efficacy of the proposed PTA in modeling local structures, visualizations of attention weights from both standard attention and PTA on the 3DMatch dataset are depicted in Fig. \ref{attention}. The PTA progressively focuses attention weights on crucial local features, whereas the standard attention considers numerous irrelevant points. These visualizations illustrate PTA's ability to concentrate on crucial local structures, thereby enabling our approach to extract rich local features effectively. Furthermore, the examples of cross-attention highlight PTA's capability to integrate information from feature-rich regions proximal to corresponding points, facilitating the network in accurately localizing target locations. Therefore, the proposed PTT can extract abundant local and global information and accurately predict correspondences.}

\subsection{Ablation Studies} \label{Ablation section}
To analyze the effectiveness of the proposed PTT, ablation studies are carried out on the 3DMatch and 3DLoMatch benchmarks, comparing it with its variants. The results are shown in Table \ref{Ablation_table}.

\textbf{Comparison with the standard transformer}. PTT$_{ST}$ is 
obtained by replacing the PTA module with the standard attention mechanism \cite{vaswani2017attention}, resulting in lower $\rm RR$ values on both the 3DMatch and 3DLoMatch benchmarks, with values of 92.3\% and 66.9\%, respectively. These results indicate that PTA enhances the local structure modeling capability, enabling the extraction of rich local information and yielding improved registration accuracy.

\textbf{Learned attention sparsity}.
\textcolor{black}{PTT$_{fixed}$ utilizes a fixed pattern for self-attention only, as fixed patterns are challenging to apply for cross-attention. In particular, PTT$_{fixed}$ specifies the attended regions as the child points of the $\mathcal{S}$ nearest coarse points instead of the relevant points, causing the $\rm RR$ values on 3DMatch and 3DLoMatch to drop to 93.1\% and 70.4\%, respectively. PTT$_{fixed\_shift}$ additionally utilizes a window shifting operation \cite{liu2021swin}, but even then, the $\rm RR$ values decrease to 93.3\% and 69.0\%. These results demonstrate that our learned attention sparsity enhances the local structure modeling capability by enabling each query to be evaluated using only high-relevance points.}

\textbf{Coarse feature guidance}. PTT$_{w/o\,  coarse\_info}$ computes the attention without the guidance of the coarse features, resulting in decreased $\rm RR$ values on 3DMatch and 3DLoMatch (91.2\% and 64.4\%, respectively). These results demonstrate that the guidance provided by the coarse features enhances the local feature extraction. 

\textbf{Feature pooling}. PTT$_{equal\_pool}$ directly pools the features from the dense layers without feature recalibration. Accordingly, the $\rm RR$ values on 3DMatch and 3DLoMatch decrease. This indicates that adaptive recalibration enhances the representative power of the features. { PTT$_{multi\_scale\_PE}$ introduces the corresponding positional encoding in each layer of the feature trees, which also leads to performance degradation.}

\textbf{Shared parameters}. PTT$_{w/o\, shared}$ leverages independent parameters rather than shared parameters in PTA, resulting in a decrease in performance. This demonstrates that parameter sharing not only reduces the number of model parameters but also enhances the generalization ability. 

\textbf{Lateral connections}. PTT$_{lateral\_connect}$ additionally introduces lateral connections \cite{lin2017feature} to aggregate the features extracted from each layer in the PTA, and the accuracy on both 3DMatch and 3DLoMatch shows a significant decline.

\begin{figure*}[!t]
\setlength{\abovecaptionskip}{-0.15cm}
    \centering
    \includegraphics[width=18cm]{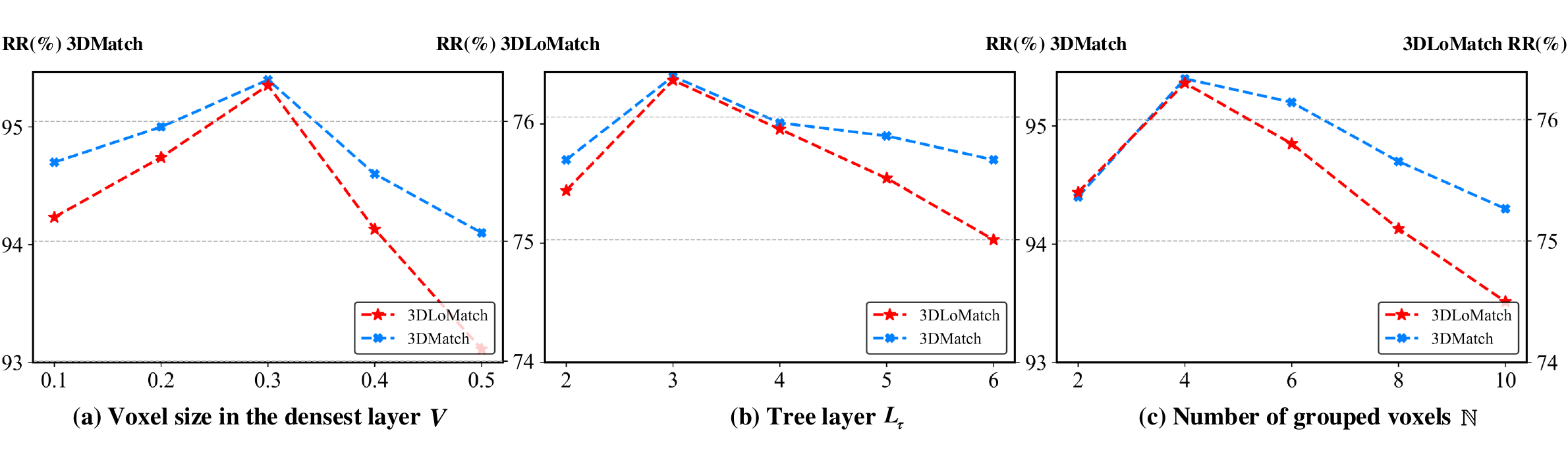}
    \caption{Ablation study results of different tree structures. Registration recall metrics, considering voxel size, tree layer, and the number of grouped voxels, are provided for the 3DMatch and 3DLoMatch benchmarks.}
    \label{tree}
    \end{figure*}

\textbf{Tree structures}.
\textcolor{black}{The configuration of tree structures depends on both the count of tree layers ($\bm{L}_{\tau}$) and the voxel sizes across various layers, with the voxel sizes being dictated by the voxel size in the densest layer ($\bm{V}$) and the number of grouped voxels ($\mathbb{N}$). 
As shown in Fig. \ref{tree}, our approach exhibits robustness to different tree structures. Favorable performance can be achieved by simply configuring $\bm{V}$ to match the voxel distance $m$ utilized in the final downsampling layer of the KPConv network ($m=0.2$ for 3DMatch), and setting $\mathbb{N}$ to 8, similar to the octree. Additional performance enhancements can be pursued through grid searches. Experimental results depict that setting parameters to excessively large or small values leads to a performance decline.  Large voxel sizes ($\bm{V}$ and $\mathbb{N}$) impede fine-grained region specification, while small voxel sizes lead to limited information within each coarse point. Shallow structures cause coarse layers to encompass numerous irrelevant regions, whereas deep structures hinder training convergence.}

\begin{table}
        \center
        \small
        \caption{Ablation results on the 3DMatch and 3DLoMatch benchmarks concerning the effects of the different model components. The RR is given in $\%$, the RRE in $^{\circ}$, and the RTE in $m$.}
        \label{Ablation_table}
        \begin{tabular}{l@{\hspace{6pt}}|c@{\hspace{8pt}}c@{\hspace{8pt}}c@{\hspace{8pt}}|c@{\hspace{8pt}}c@{\hspace{8pt}}c@{\hspace{8pt}}}
            \toprule[1.5pt]
            \multirow{2}*{\small Method}  & \multicolumn{3}{c|}{\small 3DMatch}
            & \multicolumn{3}{c}{\small 3DLoMatch}
            \\
            ~ & {\footnotesize {$\rm RR$}} &{\footnotesize {$\rm RRE$}} & {\footnotesize ${\rm RTE}$} & {\footnotesize {$\rm RR$}}&{\footnotesize {$\rm RRE$}} &{\footnotesize ${\rm RTE}$}  \\
            \midrule[1pt]
            PTT$_{ST}$& 92.3 & 1.53 &0.048 &66.9 &2.67 &0.078\\
            PTT$_{fixed}$& 93.1 & 1.61 &0.047 &70.4 &2.77 &0.080\\
            PTT$_{fixed\_shift}$& 93.3 & 1.65 &0.053 &69.0 &2.83 &0.084\\
            PTT$_{w/o\,  coarse\_info}$& 91.2 & 1.75& 0.054& 64.4& 3.45 & 0.098\\
            PTT$_{equal\_pool}$& 94.1 & 1.55 &0.048 &71.7 &{2.64} &0.078\\
            PTT$_{mult\_scale\_PE}$& {94.5} & {1.42} &{0.045} &{73.6} &2.65 &{0.074}\\
            PTT$_{w/o\, shared}$& 94.0 & 1.45 &{0.045} & 73.1 & 2.66 &0.076\\
            PTT$_{lateral\_connect}$& 92.1  & 1.55 & 0.049 & 67.3 & 2.86 &0.079 \\
            \midrule[0.5pt]
            PTT & {\textbf{\textcolor{black}{95.4}}}& {\textbf{\textcolor{black}{1.49}}}& {\textbf{\textcolor{black}{0.043}}} &{\textbf{\textcolor{black}{76.3}}}&  {\textbf{\textcolor{black}{2.26}}} &
            {\textbf{\textcolor{black}{0.067}}}\\
            \bottomrule[1.5pt]
        \end{tabular}
    \end{table}

\section{Conclusion}
In this work, a novel transformer-based network named the PTT is proposed. This approach can extract abundant local and global information while maintaining linear computational complexity. Our approach builds coarse-to-dense feature trees, and the proposed PTA module follows tree structures to progressively narrow the attended regions and structurize point clouds. Specifically, the coarse layers adaptively specify the attended regions in the dense layers based on their attention scores, and the extracted coarse features are incorporated into the dense layers to guide the feature extraction process, thereby enabling our method to focus on important local structures and facilitating local feature extraction. Importantly, the learned attention sparsity enables PTA to be used for both self-attention and cross-attention.
Extensive experiments conducted on the 3DMatch, ModelNet40, and KITTI datasets demonstrate that PTT achieves SOTA performance. 

{
\bibliographystyle{unsrt}

\bibliography{egbib}
}

\end{document}